\documentclass[journal,article,accept,moreauthors]{Definitions/mdpi} 

\DeclareUnicodeCharacter{02BE}{\textquotesingle}
\DeclareUnicodeCharacter{1E93}{\d{z}}
\DeclareUnicodeCharacter{02BF}{\textquotesingle}
\DeclareUnicodeCharacter{1E6C}{\d{T}}
\DeclareUnicodeCharacter{1E62}{\d{S}}
\DeclareUnicodeCharacter{1E24}{\d{H}}
\DeclareUnicodeCharacter{0331}{\_}
\DeclareUnicodeCharacter{1E95}{\b{z}}
\DeclareUnicodeCharacter{1E92}{\d{Z}}

\usepackage{algorithm}
\usepackage{algpseudocode}
\usepackage[utf8]{inputenc}
\usepackage[arabic,english]{babel}
\usepackage{tipa}
\usepackage{longtable}
\usepackage{amsmath}

\firstpage{1} 
\pubvolume{1}
\issuenum{1}
\articlenumber{1}
\pubyear{2025}
\copyrightyear{2025}
\datereceived{ } 
\daterevised{ } % Comment out if no revised date
\dateaccepted{ } 
\datepublished{ } 
\hreflink{https://doi.org/} % If needed use \linebreak
\pdfoutput=1 % Uncommented for upload to arXiv.org
\Title{PARSI: Persian Authorship Recognition via Stylometric Integration}

\TitleCitation{Title}

\Author{Kourosh Shahnazari $^{1,\dagger}$ \orcidA{}, Mohammadali Keshtparvar  $^{2,\dagger}$ \orcidC{}, Seyed Moein Ayyoubzadeh  $^{1,\dagger}$ \orcidB{}}

\AuthorNames{}

\isAPAStyle{%
       \AuthorCitation{Lastname, F., Lastname, F., \& Lastname, F.}
         }{%
        \isChicagoStyle{%
        \AuthorCitation{Lastname, Firstname, Firstname Lastname, and Firstname Lastname.}
        }{
        \AuthorCitation{Lastname, F.; Lastname, F.; Lastname, F.}
        }
}

\address{$^{1}$ \quad Sharif University of Technology\\$^{2}$ \quad Amirkabir University of Technology
}

\corres{}

\firstnote{These authors contributed equally to this work.}
\abstract{
Persian classical poetry, with its intricate combination of linguistic, stylistic, and metrical elements, poses significant challenges for computational authorship attribution. We present this study as an in-depth and flexible framework to identify authorship of Persian poetry, focusing on 67 prominent classical poets. Our approach employs a multi-input neural architecture combining a transformer-based language encoder with carefully designed features specifically aimed at capturing the semantic, stylometric, and metrical richness in Persian poetry. The feature space includes 100-dimensional Word2Vec embeddings, seven stylometric metrics (punctuation density, word count, etc.), and categorical representations of poetry form and meter classification. We created a large corpus of 647,653 verses (beyts) extracted from the Ganjoor digital collection, maintaining data validity using stringent preprocessing, author validation, and partitioning at the level of individual poems to ensure data non-overlap. Our empirical evaluation considers several evaluation methods, including verse-level classification and aggregation at the level of individual poems using majority voting and weighted voting, which demonstrates the latter method achieving better aggregate performance, reaching 71\% accuracy. We also investigate the application of threshold-based decision filtering, which allows the model to produce only highly confident outputs, to reach as much as 97\% accuracy at a threshold of 0.9, with the cost of reduced coverage. Our research reinforces the need to incorporate deep representational forms with domain-aware literature features to improve authorship attribution resilience. The findings highlight the utility of our approach to large-scale automated classification and its ability to advance humanistic inquiry in stylistic analysis, authorship disputes, and the wide range of computational studies of literature. The following research lays the groundwork for future research on multilingual and cross-cultural authorship attribution, stylistic change, and generative modeling in Persian and other classical forms of poetry.
}


\fancypagestyle{plain}{
  \fancyhf{} % Clear all header and footer fields
}

\begin{document}

%%%%%%%%%%%%%%%%%%%%%%%%%%%%%%%%%%%%%%%%%%

\section{Introduction}

\subsection{Background and Significance}

Persian classical poetry is a central element of the literary and cultural heritage in the world, a tradition that is well over a thousand years long. It is not only a means of art but a hardy storehouse of collective memory, a historical record, a probe into philosophy, and a meditation. Works written by great masters like \textit{Rumi}, \textit{Hafez}, \textit{Saadi}, \textit{Ferdowsi}, \textit{Khayyam}, and \textit{Jami} have echoed through the ages, enthusing listeners from generation to generation, while at the same time transgressing borders of geopolitics and languages.

The richness of Persian poetry is not only in thematic variety but also in complex formal features. Traditionally, poets worked with carefully structured forms and metres, such as Ghazal, Rubaʿi, Masnavi, and Qasida, where strict rules applied in terms of rhyme, rhythm, and rhetorical decoration \cite{hamidi2011meter,plechavc2021versification,salami2021recurrent}. Such formal features are not merely ornamental but are inherently allied to sense, often used to evoke spiritual equilibrium, emotional subtlety, or philosophical complexity. In addition, Persian poetry often makes use of symbols, allusion, allegory, and subtlety of words, concentrating a single line into a variety of meanings. This blend of style complexity and sense depth makes traditional Persian poetry a hard but rewarding field of computational study.

Over many centuries, those literary works relied on the labours of scholars, scribes, and oral reciters that preserved the collection of poems through memorization, commentaries, and manuscript reproduction. While such a traditional approach to preservation is rich in cultural values, it also created issues with orthographic inconsistencies, uncertainties regarding authorship, and stylistic similarities among writers \cite{rezaei2017stylometric,farahmandpour2015study}. All of those are still significant impediments in determining the authorship of poems, especially those published anonymously or differentiating authorship in different copies of manuscripts.

The development of digital technologies in parallel with the spread of open-access repositories revolutionized the availability and analytical power of classical literature extensively. A remarkable example of this development is the formation of \textit{Ganjoor}, a massive public repository of Persian poems. Ganjoor makes complimentary access to a broad set of verses available, while complementing each entry with contextual metadata such as the name of the poet, the poem's title, form of the verse, and metrical pattern \cite{ganjoor}. This effort towards digitalization moved Persian poetry from a static text form to a dynamic data set, thus opening ways to massive textual analysis, pattern discovery, and machine learning methods.

In spite of the greater availability of Persian poetry, it is still comparatively underrepresented in the field of computational literary analysis compared to corpora in English, in Chinese, or even in Arabic. Most current natural language processing (NLP) tools and datasets are insufficient to meet the particular demands of Persian poetic discourse. Standard methods like tokenization, lemmatization, and parsing often overlook the grammatical and rhetorical subtleties characteristic of classical Persian \cite{hamidi2011meter,salami2021recurrent}. In addition, the line-based structure of Persian poetry—where a single idea is often contained in a single line (or beyt)—requires customized methods different from document-oriented or sentence-oriented methods found in NLP.

Automated poet recognition in this framework is of significant academic and practical value. In the first place, it provides a scientific methodology whereby computational stylistics can quantify the features of poem styles, theme orientation, and the typical linguistic features of different poets. In addition, it is a decision-making tool for literary historians in order to define disputes in respect of authorship and to determine essential patterns in relation to influence, imitation, and divergence.

The value of this research goes beyond the borders of Persian studies. It enriches the broader field of Digital Humanities by showing the feasibility of advanced machine learning methods in culturally unique literary spaces. Through addressing the inherent linguistic, compositional, and cultural challenges in relation to Persian poetry, this research not only contributes to the available resources of literary analysis but also provides a model for similar research in less represented languages and literary cultures.

We outline in the sections that follow our methodology for creating a poet identification system based on the Ganjoor corpus, describing the multimodal features we obtain from poems and a hybrid model of classification that combines contextual embeddings with traditional style and structure features. Our goal is to blend traditional literary analysis with modern computational methods and thus offer a scalable, understandable, and culturally aware solution to the problem of determining authorship in poetry.

\subsection{Digitization and the Role of Ganjoor}

The Ganjoor project has had a profound impact on the study of Persian literature. As a carefully curated and open repository, Ganjoor is a critical resource for technologists working on the computational analysis of pre-modern Persian poetry and also for scholars undertaking in-depth literary analysis. Having an archive of tens of thousands of lines, a rich collection of machine-readable and standardized poems, together with careful annotations where necessary, Ganjoor is an irreplaceable resource for in-depth literary study \cite{ganjoor}.

What distinguishes Ganjoor from previous approaches to poetic archiving is its conscious emphasis on both structural features and related metadata. Every part of a poem is carefully tagged with detailed information, which includes the poet's name, the particular work or collection it belongs to, its particular form (Ghazal or Masnavi, for example), and the meter used. This systematic approach not only allows for traditional literary analysis but also makes automated processing via modern computational tools possible. Researchers can thus systematically break down verses by genre, meter, or writer, compare styles to each other, and even build annotated datasets suitable for training machine-learning models.

In addition, efforts made by Ganjoor in normalization and encoding significantly reduce the unnecessary noise usually found in digitized literary archives. This yields purified text that has often been reviewed and revised by literary experts and volunteers, thereby increasing the reliability of future natural language processing uses. In addition, by incorporating typographical conventions unique to Persian—e.g., zero-width non-joiners (ZWNJ)—Ganjoor successfully meets the demands of modern Persian-language models.

Ganjoor data contains a range of time-based, genre-based, and geographic-based features, it presents a unique opportunity to study linguistic development, poetic influence, and style evolution in an integrated context.

In our study's context, Ganjoor is not only a repository of data but, most importantly, a building block. Its range of languages, together with its structure-based annotations and broad scope of material, are essential in building rich models capable of detecting fine-grained variations in poetical voice and characteristic authorial qualities. Lacking this resource, creating an adequate and generalizable means of distinguishing between poets in Persian poetry would be much more difficult, if not impossible.

Advances in digital humanities created space for tools like Ganjoor that illustrate the intersection of open data, cultural preservation, and technological progress in a bid to protect and expand literary heritage. Such tools revolutionize the work of scholars in re-examining traditional works of literature using modern approaches and enabling the creation of new kinds of questions—queries that move past disciplinary boundaries and combine computational analysis with humanistic inquiry.

\subsection{Challenges in Computational Analysis}

In spite of the availability of digital corpora like Ganjoor and advances in natural language processing technologies, computer analysis of traditional Persian poetry is faced with a unique set of challenges arising from the richly complex nature of the poem form, linguistic subtleties, and limits of available resources. In addition to complicating algorithmic analysis, such challenges also require the creation of novel modeling paradigms capable of balancing computational precision with literary insight.

One major problem in the analysis of Persian poetry lies in the dense and complex structure of the beyt, the building block of poetry. Every beyt is relatively independent, holding metaphorical value, sense ambiguity, and semantic depth in just two hemistichs, unlike longer texts or paragraphs. This high concentration of meanings poses challenges to computational models, especially those learned using prose data, to capture the entire range of expressions inherent in any verse adequately \cite{salami2021recurrent}. Conventional natural language processing tools typically make the assumption of long contexts and often struggle in creating senseful embeddings of such short and symbolically dense pieces \cite{hamidi2011meter}.

One major problem involves the widespread stylistic convergences that appear amongst the poets. A commitment to well-tried formal conventions—most notably in relation to meter, rhyme, and genre—of the Persian literary tradition often results in situations where multiple poems share similar stylistic features \cite{rezaei2017stylometric,plechavc2021versification}. Such similarity is most visible amongst poets sharing the same chronological context or literary movement. Untangling those closely interconnected styles requires analytical tools that are powerful enough to recognize subtle signs in vocabulary, sentence structure, imagery, and even phonic construction.

Linguistic diversity brings a degree of complexity to computational approaches. Persian poetry often utilizes archaic words, words found in regional dialects, Arabic loan words, and obsolescent syntactic constructions inadequately represented in modern Persian corpora or in tokenization models \cite{rahgozar2016bilingual,hamidi2011meter}. In addition, having numerous morphological variants, no capitalization, and recurring uses of metaphor and ellipsis make it difficult to perform part-of-speech tagging, named entity detection, and syntactic analysis.

Structurally, orthographic irregularities—variations in spelling, inconsistent use of diacritics, and typographic deviations—are the sources of disturbances in text normalization processes, embedding alignment, and tokenization \cite{hamidi2011meter,salami2021recurrent}. While some challenges are mitigated in scrupulously prepared datasets such as Ganjoor, they are persistent challenges to extensive analytic and generalization endeavors. In addition, traditional Persian poetry has a level of visual organization and symbol-based structuring (e.g., the symmetry of verses and the beauty of calligraphy) incongruent with typical NLP paradigms but necessary for a rich analysis of poem texts.

Another serious hurdle is the scarcity of available annotated datasets and benchmark tasks specifically targeting Persian poetry \cite{shahnazari2025nazmnetworkanalysiszonal,salami2021recurrent}. Unlike the English-language natural language processing domain, which has a multitude of large labeled corpora for a wide range of subtasks, Persian poetry lacks such resources. This absence hinders the use of supervised learning methods, worsens data imbalance issues, and forces reliance on semi-supervised or transfer learning methods that can have poor generalization.

Finally, the uncertainty of authorship attribution poses great conceptual and practical challenges. Many years of manuscript circulation, coupled with editorial revisions and stylistic imitation, have left us with a body of poems of doubtful or ambiguous authorship \cite{farahmandpour2015study,rezaei2017stylometric}. A computational model, therefore, needs to capture this uncertainty, express probabilistic confidence, and facilitate interpretability to be maximally useful for scholars. In addition, the model should exhibit robustness to overfitting to irrelevant lexical cues that do not necessarily reflect the author's voice.

Together, these challenges carry a particular significance of the need for hybrid approaches that combine neural representations with expert-annotated metadata, carefully engineered features, and strict evaluation procedures \cite{salami2021recurrent,plechavc2021versification}. Only through such integrated approaches can we begin computationally to recreate the richness, depth, and nuances characteristic of the tradition of Persian poetry.

\subsection{Recent Advances in NLP}

The last decade witnessed a revolutionary transformation in the world of Natural Language Processing (NLP), spurred mostly by innovations and improvements in the technologies of deep learning, especially those based on the transformer model. First proposed by Vaswani et al. \cite{vaswani2017attention} in their landmark work "Attention is All You Need," transformers later grew to be the go-to framework for most advanced language models in use today, revolutionizing the way machines understand, generate, and interact with human language.

Some of the most prominent transformer-based models include BERT (Bidirectional Encoder Representations from Transformers), GPT (Generative Pre-trained Transformer), RoBERTa, XLNet, and T5. One of the most striking features of such models is their ability to learn rich contextual representations of language through self-supervised pretraining on massive text corpora followed by fine-tuning towards a target downstream task. In contrast to earlier models, which were largely based on hand-engineered features or recurrent architectures, transformers leverage multi-head self-attention mechanisms to represent long-distance dependencies and syntactic-semantic relations in text \cite{vaswani2017attention,alqurashi2025bert,qarah2024arapoembert}.

This shift has dramatically increased the effectiveness in a broad spectrum of tasks, particularly those like text classification, question answering, named entity recognition, summarization, and machine translation \cite{alqurashi2025bert,qarah2024arapoembert}. Of direct utility in our research, transformers have made it possible to identify complex stylistic and semantic trends, thus making them ideally suited to literary analysis and authorship attribution \cite{salami2021recurrent}.

Traditionally, support for low-resource languages like Persian has been largely in the form of generic models. Recently, however, with initiatives in the Persian NLP domain, models focused on particular domains and developed specifically for low-resource languages like Persian are becoming more available \cite{salami2021recurrent,rahgozar2016bilingual}. These models are based on corpora of the Persian language, including news texts, Wikipedia articles, and social user-generated data, thus enabling the inclusion of linguistic features typical of the Persian language, e.g., morphology, compounding, and script-related aspects like zero-width non-joiners (ZWNJ).

One of the main advantages of these models is their suitability for poetic literature. Even though they were not specifically trained on poetry, their performance on tokenization, handling of rare vocabulary, and acceptance of diverse syntactic structures make them exceptionally well-suited to the intricacies of classical Persian poetry \cite{salami2021recurrent}. Additionally, the availability of APIs and open-source environments makes it convenient to integrate these models into larger classification systems for researchers.

Methodologically, transformer-based models enable the inclusion of multiple feature modalities. As an example, the representation of the [CLS] token in a RoBERTa encoder can be composited with expert-designed features—e.g., stylometric features, representations of meter, and semantic embeddings—yielding a single feature vector \cite{salami2021recurrent}. This scenario is especially beneficial in multi-input models where a combination of deep-learning representations with expert knowledge is integrated, yielding models that are both precise and transparent.

In the suggested framework, a version of the RoBERTa model is used as the base encoder for verse-level text data, thus taking advantage of its inherent strengths. This encoder is able to efficiently capture the contextual relationships and the underlying semantic richness of every line of the poem, and we augment this by adding features extracted from the quantitative and structural aspects of the poem \cite{salami2021recurrent}. These include average hemistich length, word frequency distributions, meter types, form, and semantic vectors using Word2Vec. The combination of transformer-based embeddings with additional features enables our model to also pick up on subtle differences in poetic style and the author's unique voice, which would be difficult to detect if either modality is used alone.

Beyond their modeling capabilities, transformer-based approaches also bolster the analysis of results and outcome interpretability. Analytics tools such as attention visualization, embedding clustering, and layer-wise analysis provide tremendous insight into how the model is interpreting and distinguishing between different poets' works. These approaches allow researchers to carry out not just quantitative analysis of model effectiveness but also qualitative analysis of the linguistic underpinnings of model predictions, opening the door to fresh opportunities for collaboration between computer scientists and literary scholars.

In sum, the advent of transformer-based natural language processing brings a revolutionary period within the field of computational literary analysis \cite{vaswani2017attention}. It is possible, through the repurposing of such advanced technologies to suit the unique nature of Persian poems, to create models that pay homage to the richness of the original poems while offering scalable and efficient remedies to issues like identification of the poet \cite{alqurashi2025bert,salami2021recurrent}. As transformer models like Longformer, BigBird, and GPT-4 continue to advance—expanding the range of tasks that can be applied—the prospect of machines meaningfully interacting with the poetical works that define humankind's cultural inheritance will likewise increase.

\subsection{Our Approach and Objectives}

The main contribution of this work is to find a computational approach that is able to identify a particular poem's author, even in the case of restricted textual and stylistic features. To this end, we propose a hybrid model that combines the powerful representational capabilities of transformer-based models with the fine-grained, domain-sensitive features based on carefully designed literary features \cite{salami2021recurrent,plechavc2021versification}.

The center of our framework is defined by a multi-input structure capable of ingesting a variety of sources of information. Every beyt is processed through a model based on RoBERTa. This methodology produces a dense contextual representation for every beyt in an effort to capture the complex interactions found in lexical usage, semantic composition, and rhetorical structure \cite{vaswani2017attention,alqurashi2025bert}.

Although this broad representation is vital in capturing the semantic richness of poetics, we complement it with additional features that describe the formal and stylistic features of poems. These features include:

\begin{itemize}
\item \textbf{Metrical Information:} In one-hot vector form, the meters enable differentiating among those poets showing preference towards different rhythmic schemes \cite{hamidi2011meter}.
\item \textbf{Poetic Form:} These poetic forms, such as the ghazal, masnavi, and rubaiyat, are distinctive indicators of stylistic directions.
\item \textbf{Stylometric Features:} These stylometric features include the average length of words, the ratio of hapax legomena, punctuation density, symmetry ratio, and hemistich length; together, these features reflect the poet's linguistic inclinations \cite{rezaei2017stylometric,salami2021recurrent}.
\item \textbf{Semantic Embeddings:} Word2Vec embeddings, pre-trained on classical Persian poetry, capture distributional semantics that complement RoBERTa's contextual encodings \cite{salami2021recurrent}.
\end{itemize}

The entire feature stream is fed in and then passed into a fully connected classification model. The model is a multi-layer perceptron with non-linear activation and dropout regularization, and is aimed at minimizing the cross-entropy loss of predicted poet labels compared to actual labels \cite{salami2021recurrent}.

To ensure the proper generalization of our model, we employ a data splitting strategy at the poem level. This strategy prevents the leakage of information from the train to evaluation datasets, a problem specifically applicable in the domain of poetry, where multiple verses of the same poem could otherwise be split in different subsets \cite{salami2021recurrent}. Model evaluation is performed at both verse and poem levels using majority voting and probability-based combination to obtain definitive poem-level author labels from verse-level predictions.

Our approach is not just aimed at maximum performance but also at improved interpretability and flexibility. By combining derived and hand-engineered features, we provide a basis in literary theory while allowing for empirical optimization \cite{plechavc2021versification,salami2021recurrent}. In addition, this framework allows for modular experimentation, e.g., testing different embeddings, adding prosodic features, or using graph-based representations of poetic form.

In conclusion, our objectives are threefold:

\begin{itemize}
\item To construct a high-quality, feature-rich dataset of Persian poems suitable for machine learning \cite{shahnazari2025nazmnetworkanalysiszonal};
\item To design a flexible, interpretable hybrid classification model that integrates contextual embeddings with literary metadata \cite{salami2021recurrent};
\item To empirically validate the model's effectiveness using rigorous verse-level and poem-level evaluation schemes \cite{salami2021recurrent}.
\end{itemize}

With this methodology applied, our aim is to refine the classification of the Persian poets in addition to simultaneously providing results applicable in the interdisciplinary context of literary analysis and computational linguistics.

\section{Related Work}
\label{sec:related-work}

Authorship attribution in classical poetry, particularly in Arabic, Persian, and Urdu languages, is a multifaceted research domain that blends literary analysis with advanced computational techniques. This section extensively reviews prior works, emphasizing methodologies, feature sets, datasets, models, and the unique challenges inherent in these languages and poetic forms.

\subsection{Arabic Poetry Authorship Attribution}

Early studies by Ahmed et al. laid the groundwork for authorship attribution in Arabic poetry\cite{ahmed2015authorship, ahmed2016authorship, ahmed2017machine, ahmed2019arabic}. Their research leveraged classical machine learning approaches such as Naïve Bayes (NB), Support Vector Machines (SVM), and Sequential Minimal Optimization (SMO) trained on features like rhyme, meter, sentence length, word length, and first word in sentence. These studies highlighted the effectiveness of textual features in distinguishing authors, achieving high accuracies (e.g., 96.96\% and 98.96\%). However, their reliance on shallow features limited their ability to capture deeper syntactic and semantic nuances inherent in Arabic poetry’s rich structure.

To address these limitations, Boukhaled and Amine introduced a machine learning framework that emphasized syntactic style markers over purely lexical or structural features\cite{boukhaled2022machine}. Their work demonstrated that syntactic-based features could outperform traditional lexical-based methods, particularly in the context of classical Arabic texts, by capturing stylistic subtleties often overlooked in earlier models.

Building on the success of deep learning in NLP, Alqurashi et al. developed a BERT-based ensemble model trained on the entire Classical Arabic Poetry corpus, incorporating Embedded Topic Modeling (ETM) for topic annotation\cite{alqurashi2025bert}. This approach achieved F1 scores ranging from 0.97 to 1.0, highlighting the power of transformer architectures in capturing both topical and stylistic dimensions of poetry. Notably, their model handled historical misattribution cases, providing consistent results with established literary scholarship, thus bridging computational and humanistic analyses.

Meter detection, a critical feature in Arabic poetry, has been tackled from multiple angles. Berkani et al. proposed the Arabic Meters Identification System (AMIS), which combines phonological verse preparation, pattern matching, and similarity measures\cite{berkani2020pattern}. Their system achieved a remarkable 99.3\% precision, underscoring the importance of phonological processing in Arabic metrical analysis. Complementing this, Mutawa et al. explored deep learning architectures—including LSTM, GRU, and Bi-LSTM—applied to full-verse and half-verse data\cite{mutawa2025determining}. By preserving diacritics and using character-level encodings, they reported accuracies of 97.53\% and 95.23\% respectively, thus demonstrating the potential of neural models to handle the intricacies of Arabic meter.

Yousef et al. extended RNN-based models to Arabic and English poems, focusing on the direct classification of poem meters from plain text without relying on handcrafted features\cite{yousef2019learning}. Their approach signaled a shift towards end-to-end learning frameworks capable of capturing rhythm and meter implicitly from raw textual inputs.

The linguistic underpinnings of meter were further investigated by Scott, who critically evaluated al-Khalil’s classical system and compared it with modern linguistic theories such as binarity and prosodic constraints\cite{scott2010pegs}. This work emphasized the need to align computational models with the linguistic theories that govern classical Arabic poetry, highlighting opportunities for more linguistically informed computational approaches.

Furthering the integration of deep learning, Qarah introduced AraPoemBERT, a transformer-based language model pre-trained on a large corpus of Arabic poetry\cite{qarah2024arapoembert}. The model achieved state-of-the-art results in multiple tasks—including rhyme classification, sentiment analysis, and meter detection—demonstrating the potential of transformer architectures to capture the multifaceted structure of Arabic poetry.

\subsection{Urdu Poetry Authorship Attribution}

Urdu poetry, particularly the ghazal form, poses unique challenges due to its rich literary tradition and stylistic diversity. Rao and Ahmed tackled the attribution task using SVMs with unigram and bigram features, achieving 88.7\% accuracy on couplet classification for five poets\cite{rao2021poet}. This approach emphasized the efficacy of simple lexical features, but also highlighted the challenges in capturing deeper stylistic elements with limited context.

Tariq et al. expanded on this work by incorporating multiple classifiers—SVM, Decision Trees, Random Forests, Naïve Bayes, and KNN—combined with feature selection techniques like chi-square and L1-based selection on a dataset of 4000 ghazals\cite{tariq2019identification}. Their study reinforced the importance of combining content and style features for accurate classification.

Pushing the envelope further, Siddiqui et al. leveraged deep learning and transformer-based models (BERT and roBERTa) to attribute authorship across 15 poets using 17,609 couplets\cite{siddiqui2023poet}. Their results—achieving 80\% accuracy—demonstrated the superiority of transformer architectures in capturing complex stylistic patterns and contextual dependencies in Urdu poetry.

\subsection{Persian Poetry Authorship Attribution}

In the context of Persian poetry, Rezaei and Kashanian analyzed word length and richness to differentiate poets such as Attar, Molavi, and Nezami. Although these features offered initial insights, they underscored the need for more sophisticated stylistic and metrical analyses\cite{rezaei2017stylometric}.

Hamidi and Razzazi proposed a meter classification system for spoken Persian poetry\cite{hamidi2011meter}. Their system utilized pitch frequency and syllable duration features fed into an SVM classifier, achieving 91\% accuracy across 12 Persian meter styles. This approach bridged the gap between acoustic and textual analysis, illustrating the multimodal nature of Persian poetry research.

Rahgozar and Inkpen developed a bilingual corpus of Hafez’s ghazals, employing SVMs with Latent Dirichlet Allocation (LDA) and Principal Component Analysis (PCA) to classify poems chronologically\cite{rahgozar2016bilingual}. Their work highlighted the interplay between thematic content and temporal evolution in Persian poetry.

Farahmandpour and Nikmehr investigated intelligent authorship attribution using lexical, syntactic, semantic, and application-specific features, applying KNN and Linear Discriminant Analysis \cite{farahmandpour2015study}. Their study compared classifier performance and emphasized the need for feature diversity.

A more recent contribution by Shahnazari et al. introduced NAZM, a network analysis model that maps semantic, stylistic, thematic, and metrical similarities between poets. By combining features from multiple levels of analysis, NAZM provided a holistic view of Persian poetic influence\cite{shahnazari2025nazmnetworkanalysiszonal}.

\subsection{General and Cross-Language Authorship Attribution}

Cross-linguistic studies include Elayidom et al. who addressed authorship attribution by focusing on texts with undecided authorship and developed a computational stylometric approach to resolve issues such as uncertain authorship, recognition of unknown texts, and plagiarism detection\cite{elayidom2013text}. They emphasized the utility of statistical methods—including word length, sentence length, and vocabulary richness—to capture stylistic differences quantitatively. Each author was assumed to have a unique, inherent writing style that could be numerically differentiated using statistical features. Their pipeline included key steps: pre-processing, feature extraction, classification, and finally, authorship attribution itself. In their experiments, they employed both fuzzy learning classifiers and support vector machines (SVMs) to perform classification tasks. While the SVM alone demonstrated superior accuracy compared to the fuzzy classifier, the combined use of both classifiers achieved even higher performance, highlighting the benefits of classifier fusion in authorship studies. This modular and comparative approach underscores the adaptability of stylometric techniques across different textual genres and languages.

Muldoon et al. surveyed modern stylometric techniques using machine learning on chat logs, underlining the adaptability of these techniques to varied textual genres\cite{muldoon2021modern}.

Plecháč explored versification features—such as rhyme and rhythm—as potential stylistic fingerprints, reinforcing the relevance of poetic structures in authorship studies\cite{plechavc2021versification}.

Agirrezabal et al. compared feature-based and neural approaches to scansion in English and Spanish poetry, highlighting the strength of Bi-LSTM+CRF models over handcrafted features\cite{agirrezabal2017comparison}.

Haider proposed multi-task learning frameworks with syllable embeddings, demonstrating inter-task dependencies in poetic rhythm analysis\cite{haider2021metrical}.

\subsection{Datasets and Corpora}

A key resource for Arabic poetry research is Al-Onazi et al. ’s “Diwan,” which represents the largest and most precisely annotated corpus of Arabic poetry to date, comprising approximately 14 million verses across 16 major categories\cite{al2025diwan}. This dataset is distinguished by its comprehensive scope and the depth of its annotations, which include detailed prosodic structures, thematic content, linguistic patterns, and poet-specific metadata. These annotations were meticulously curated through advanced data collection methods and rigorous normalization protocols, overseen by experts in Arabic prosody and poetry. The “Diwan” corpus addresses key challenges in Arabic poetry analysis, such as the language's complex metrical structures, diverse themes, and unique linguistic intricacies that often hinder conventional NLP models. By providing a robust foundation for AI-powered analysis and deep-learning-based research, “Diwan” enables a wide range of tasks including automatic poetry generation, metrical analysis, thematic classification, and plagiarism detection. Comparative evaluations against four leading Arabic poetry corpora confirm that “Diwan” outperforms existing datasets in both scope and annotation quality, solidifying its status as an indispensable resource for computational literary studies and digital humanities research in the Arabic language.

\subsection{Hybrid and Deep Learning Approaches}

Salami and Momtazi proposed a recurrent convolutional neural network (RCNN) architecture that effectively combines the strengths of RNNs and CNNs for poet identification\cite{salami2021recurrent}. While RNNs are adept at capturing temporal dependencies in poetic texts, CNNs excel at identifying local stylistic patterns, making their integration particularly well-suited to analyzing poetry. To address the limitations of standard RNNs, Salami et al. incorporated advanced recurrent units like LSTM and GRU to enhance the model's ability to learn long-range dependencies without suffering from vanishing gradients. Their model was evaluated at both the poem and verse levels, achieving substantial improvements—9\% and 4\% in f-measure, respectively—over existing CNN-based approaches. This hybrid architecture demonstrated superior performance by capturing both sequential flow and stylistic features of poems, highlighting the potential of deep learning models to preserve and analyze literary heritage in the field of authorship attribution.

\section{Methodology}

\subsection{Dataset}
\subsubsection{Source and Scope}
\label{Source and Scope}

Our dataset is constructed from the Ganjoor project (\url{https://ganjoor.net}). It is the most massive and publicly available digital collection of classical Persian poetry. It contains over 693,000 verses (\textit{beyts}) written by 74 poets, spread over a literary span of over ten centuries—ranging from the epic poems of Ferdowsi through the mystical masnavis of Rumi and the philosophical quatrains of Khayyam. A rich variety of poetic forms and metrical patterns are exhibited in the corpus, thus testifying to depth and variety inherent in the Persian literary tradition \cite{shahnazari2025nazmnetworkanalysiszonal}. Ganjoor's carefully structured layout and standardized formatting make it an extremely suitable object for computational analysis, including authorship attribution.

In order to ensure the integrity and quality of the training data, our choice was limited to poems that had unequivocally accepted and uncontested authorship. Pieces with uncertain or multiple attributions were excluded from the training and reserved solely for later evaluation. We also excluded poets whose oeuvre comprised less than 50 verses in order to prevent problems associated with class imbalance and underrepresentation in supervised learning settings.

\vspace{0.5em}
The final dataset statistics are as follows:

\begin{itemize}
    \item \textbf{Total poets:} 74
    \item \textbf{Total poems:} 60,327
    \item \textbf{Total verses:} 693,158
    \item \textbf{Poetic forms:} 20 distinct types (e.g., ghazal, qasida, masnavi, robāʿī, do-beyti)
    \item \textbf{123 canonical meter patterns}, annotated in traditional Persian \textit{aruz}.
\end{itemize}

Each poem in our collection is broken down into separate \textit{beyts}—the fundamental building block of Persian poetry, typically two hemistichs that together express a single poetic idea. This is in keeping with the precepts of classical Persian poetics and ensures that, in breaking down the data, we respect the inherent boundaries of sense and measure that are part of the verse. To enable the successful harnessing of the poetic and compositional complexities inherent in Persian poetry, we create a multi-modal feature set for every beyt. The feature representation includes a range of textual, semantic, stylistic, and formal features, chosen carefully to reflect various facets of poetic expression, all contributing collectively to the authorship classification task.

\paragraph{Textual Input.}
The first step in processing any beyt text is applying a tokenizer that obeys Persian linguistic rules, capable of handling orthographic marks like zero-width non-joiners (ZWNJ) proficiently \cite{hamidi2011meter,salami2021recurrent}. This is necessary in order accurately to segment compounding expressio

\subsubsection{Data Extraction and Preprocessing}
\label{Data Extraction and Preprocessing}

The textual data employed in this current study were collected through automated web scraping methods from the online repository Ganjoor (\url{https://ganjoor.net}), a rich collection of publicly available classical Persian poems. In order to enable systematic collection of the corpus, a customized web scraping system was developed that collected poems alongside their metadata, including poem title, genre, traditional \textit{aruz} metrical tags, poet name, and full verse text. Through this methodology, a unified and rich dataset could be developed by taking advantage of Ganjoor's structured metadata such as poem title, genre, traditional \textit{aruz} metrical indicators, poet name, and full verse text \cite{shahnazari2025nazmnetworkanalysiszonal,salami2021recurrent}.

Following collection and acquisition of the raw data, a thorough multi-phase preprocessing structure was applied to translate the text corpus into a processed, balanced, and machine-friendly form suitable for authorship classification analysis. Every phase of this structure was carefully designed to address specific issues inherent in Persian poems, such as orthographic diversity, ambiguous authorship attributions, and stylistic variety. To illustrate, the use of text normalization methods standardized characters, removed unnecessary diacritics, and normalized typographic usage, so that the data followed a uniform encoding framework \cite{hamidi2011meter,salami2021recurrent}.

Therefore, we performed the tokenization using a language-suitable tokenizer capable of handling Persian-specific orthographic constructs, particularly zero-width non-joiners (ZWNJ), which are essential for preserving morphological consistency in compound words and affixed constructions~\cite{hamidi2011meter}. Unlike standard normalization pipelines that typically remove ZWNJs to reduce textual variability, we retained them throughout our preprocessing to maintain linguistic integrity and improve downstream model performance. Tokenization enabled the construction of reliable input sequences compatible with the transformer-based models discussed in this study.

Eventually, feature extraction was applied per verse (beyt), yielding a mix of learned features and hand-designed ones—semantic embeddings, stylometric features, and categorical encodings pertaining to poem form and meter. Upon standardization, the extracted features were subsequently joined together with the tokenized text input in order to construct a unified multi-modal representation per instance. Our developed dataset not only preserves the finer details of style and structure characteristic of Persian poetry but also provides a solid foundation for the construction of accurate and understandable machine learning models.

\paragraph{1. Authorship Attribution.}
One major impediment faced in the study of historical poems relates to the uncertainties in authorship attribution. Thankfully, Ganjoor offers metadata markers in relation to each poem classifying authorship into confirmed, contested, or ambiguous. In preparing the training dataset, poems labeled as \textit{contested}, \textit{anonymous}, or \textit{ambiguous} were systematically excluded. These examples were not abandoned per se but reserved for a follow-on task highlighting the model's ability to determine ambiguous authorship through predictive analysis \cite{farahmandpour2015study}.

\paragraph{2. Verse Segmentation.}
In the context of Persian poetry, the defining feature of rhythm and sense is the \textit{beyt}, a two-part hemistich couplet. Rather than train our model using entire poems, we decided to consider each beyt an independent class example. This allows us to achieve greater accuracy in attribution and increase the size of our dataset. Verse segmentation was carried out using Ganjoor's structure tags in addition to linebreak conventions, thus ensuring the proper and consistent extraction of the beyts from the poem's original text \cite{hamidi2011meter}.

\paragraph{3. Text Normalization.}
Persian orthography poses many challenges for computational modeling, such as irregularities in Unicode encodings, overlapping features of the Arabic script, and typographic variation. In order to overcome these issues, we established an extensive normalization pipeline. This included:

\begin{itemize}
\item Replacing Arabic "yeh" with Persian "ye", and Arabic "keh" with Persian "kaf";
\item Removing all diacritics like vowel markers, and removing zero-width non-joiners that disrupt tokenization;
\item Stripping editorial markers or inline tags, like HTML span tags, from the original documents;
\item Standardizing punctuation conventions and controlling the space between tokens.
\end{itemize}
The normalization process ensured that stylistic variations would not interfere with the model, while ensuring consistent tokenization of text sequences across the whole dataset \cite{hamidi2011meter}.

\paragraph{4. Label Encoding and Feature Preparation.}
We assigned each poet a unique integer identifier using LabelEncoder from the scikit-learn library, thus creating a classification target. In addition to the textual data of the poems, we used a number of extra attributes—namely, meter structure and form of the poems. As both features were categorically based, a conversion to one-hot encoded vectors using the OneHotEncoder tool followed. This allowed the inclusion of non-textual literary features in the model framework, and this improved the precision in classification.

\paragraph{5. Dataset Splitting.}
To prevent leakage of data, it was necessary to ensure that verses from the same poems were not contained in both the train and validation or test subsets. Therefore, we split at the \textit{poem level} and not the verse level. To ensure that the entire range of poet labels remained consistent in the different splits, we applied stratified sampling methods. Through this, we were able to ensure that any subset correctly represented the entire dataset, while keeping the poems' integrity intact \cite{salami2021recurrent}.

\subsubsection{Dataset Statistics and Overview}

To provide a broad overview of the structure and nature of the data, we provide extensive statistics and exploratory analysis that highlight the size, variety, and balance of the corpus. Here, the overall numbers of poems, authors, and verses; the split of poetic types and metrical forms; alongside the features determining the development of the data in terms of authorship classification goals are outlined \cite{shahnazari2025nazmnetworkanalysiszonal,salami2021recurrent}.

\vspace{0.5em}
\noindent
\textbf{Poet Distribution.}
The collection represents poetry written by 74 poets drawn from the Persian literary canon. The distribution of poems written by each poet shows a heavy skew, reflecting historical processes like the survival of manuscripts, the popularity of the poets, and their literary influence \cite{shahnazari2025nazmnetworkanalysiszonal}.

The average poet produces approximately 815.2 poems; however, the variance is quite high amongst the poets. The most represented poet in the dataset is \textit{Mowlānā}, with a staggering total of 6320 poems. In contrast, at the lower end of the spectrum is \textit{ʿAbd al-Wāsiʿ Jibilī}, whose contribution is a mere 6 poems. A high variance is a challenge to supervised learning since the models will be biased towards those poets that are most productive.

\begin{figure}[ht]
    \centering
    \includegraphics[width=0.7\linewidth]{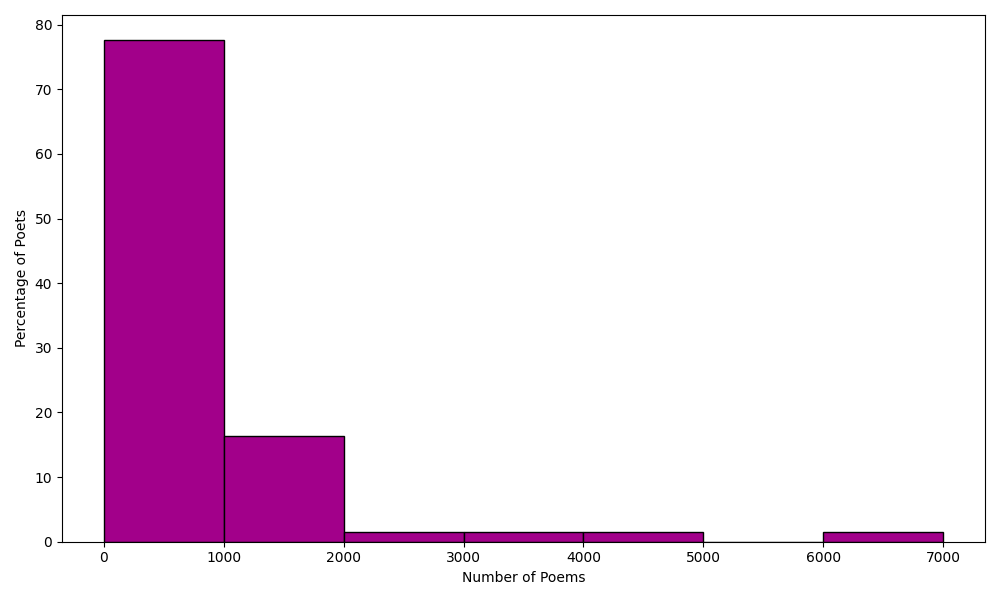}
    \caption{Distribution of Poems per Poet. The dataset is highly imbalanced, with a few prolific poets and many less-represented ones.}
    \label{fig:poet-distribution}
\end{figure}

Figure~\ref{fig:poet-distribution} illustrates this imbalance, displaying the number of poems written by established authors. This histogram is heavy-tailed, where a few authors contribute a large number of poems compared to the many authors producing relatively few works.

\noindent
\textbf{Poetic Structures.}
The entire body of work exhibits a remarkable diversity of poetical form in 20 different genres, thus showing the richness and diversity of Persian literary tradition \cite{shahnazari2025nazmnetworkanalysiszonal}. These genres serve not only to provide a framework but also to play an active role in creating thematic matter, rhetoric, and the total poem experience.

Amongst the different types contained therein, the most numerous is the \textit{ghazal}, amounting to a total of 23,597 poems—this confirms its long-standing popularity in the works of Persian literature, often tackling themes of mysticism, love, and earthly-divine relations. In contrast, the collection includes extremely rare types, such as \textit{bahr-e tawil}, featuring only 1 poem, thus highlighting the large extent of formal variety characteristic of the corpus.

\begin{figure}[ht]
\centering
\includegraphics[width=0.7\linewidth]{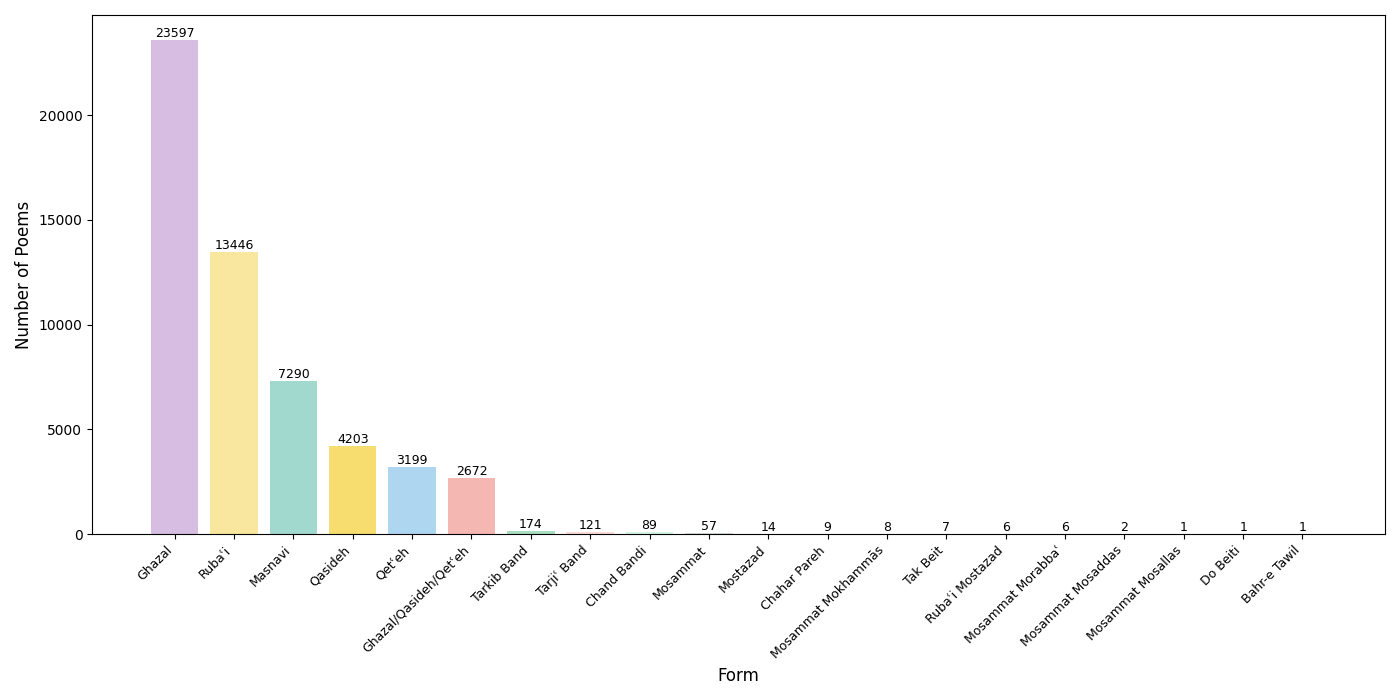}
\caption{Prevalence of Different Types of Poems in the Corpus. Ghazals are the most dominant, followed by other well-known types like masnavi, qasida, and robāʿī. Smaller groups of less common types like bahr-e tawil are also presented but with much lower rates.}
\label{fig:form-distribution}
\end{figure}

The average number of typical representations in the different kinds of poems is approximately 2745.1; however, this masks quite an uneven split between the kinds. Figure~\ref{fig:form-distribution} shows comparative occurrences of the different kinds, highlighting the discrepancies with which models have to cope when trying to identify the poet's characteristic features.

Another aspect to be taken into consideration in this dataset is the range of forms covered by various poems. Some poems demonstrate a wide range of stylistic treatment, trying their hand at many forms and meters, while others demonstrate focused expertise within a particular genre. For example, \textit{Khvāju Kermānī} presents himself as the most versatile poet in this collection, having written in a whopping 13 different forms \cite{shahnazari2025nazmnetworkanalysiszonal}. This degree of variation is a testament to his versatility and range, along with his treatment of a wide range of thematic and rhetorical genres that are characteristic of Persian literature. Conversely, \textit{Khayyām} is defined by a unique and singular poetic structure that mirrors his focused priority on the \textit{rubāʿī}—a form that dovetails perfectly with his philosophical and reflective poetry. This priority aligns with \textit{Khayyām's} historical reputation as a great master of the concise but powerful quatrain, often infused with existential themes.

The average number of different types used by a single poet is 6.0, indicating that most poets use a significant number of different types of poem. A variety of poem types makes the style of the collection richer but creates issues that are both challenging and rewarding regarding the task of authorship attribution. Models therefore need to determine poet identity on the basis of form-based features—e.g., structure, meter, and rhyme scheme—whose effect on classification decisions might otherwise be preponderant.

\vspace{0.5em}
\noindent
\textbf{Meter Patterns.}
Meter, or \textit{aruz}, is an integral element of Persian poetry, having a significant effect on the poem's rhythmic and phonic structure, together with its semantic development and possible meanings \cite{hamidi2011meter,salami2021recurrent}. Up to 123 various metrical schemes are contained in the collection, highlighting the immense richness in ancient Persian poetical tradition. To find an equilibrium between representativeness and computational feasibility, the schemes were grouped into 15 normalized classes of meter. This approach preserves the stylistic variety typical of traditional \textit{aruz} while offering an adequate number of examples in every class to serve the goals of statistical analysis and supervised machine learning.

\begin{figure}[ht]
\centering
\includegraphics[width=0.7\linewidth]{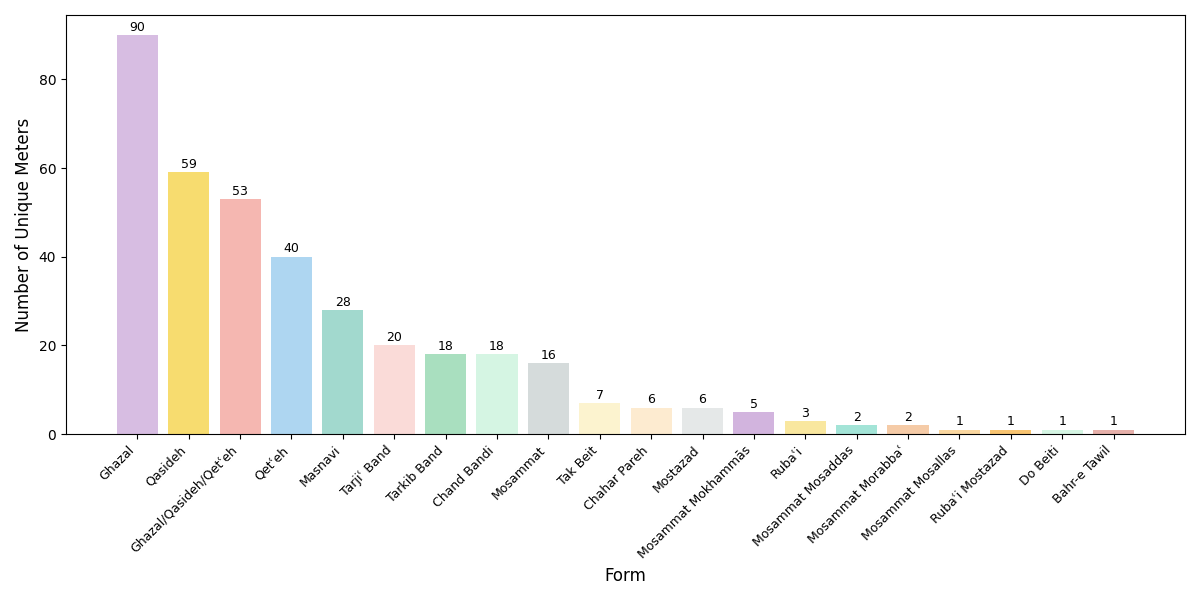}
\caption{Histogram showing the proportion of meter classes per poetic form. Ghazals dominate in meter diversity, while some forms like bahr-e tawil remain highly specialized.}
\label{fig:meter-distribution}
\end{figure}

The proportion is specifically relevant to authorship analysis since meter is not only a feature of a poet's personal style but can also be a source of ambiguity in that different writers prefer particular meters. Therefore, models should be equipped with the power to separate these effects in order to make consistent and understandable authorship predictions.

\noindent
\textbf{Verse-Level Granularity.}
With respect to the unique structure found in Persian poetry whereby each \textit{beyt} (verse) is an independent unit, our data remains at the level of separate verses \cite{hamidi2011meter,salami2021recurrent}. Such a methodology is essential in countering the wide range of stylistic variations in Persian poetry, in that discrete beyts often contain profound sets of poetical ideas, rhetorical devices, or thematic development. Through a consideration of an independent data point per verse, we enable the classification model's efficacy in recognizing fine details that would be otherwise imperceptible if representations of the poem are taken in isolation.

The collection contains a total of 647,653 verses, thus highlighting both its significant size and the broad range it offers to full statistical analysis. On average, single poems contain some 11.8 verses; however, this average masks a great degree of variance in the collection. While some poems contain a single verse, the longest poem comprises 966 verses, thus reflecting the range of diversity in structure in the poems—ranging from the brief \textit{robaʿī} form to the longer \textit{masnavis}. Median number of verses per poem is 7.0, and this means that a high percentage of the collection includes relatively short poems, a characteristic typical of the traditional brevity in poems such as \textit{ghazal} and \textit{qet'a}. The standard deviation of verses per poem is found to be 23.0, indicating a high level of distributional variability. This gap highlights the need to account for the variation in poem length in both modeling and evaluation procedures. Longer poems, for example, may provide more training samples per author, potentially skewing the learning curve of the model. By maintaining the granularity at the verse level, adaptive aggregation methods such as majority voting and weighted voting at the poem level can be used for evaluation. Such approaches enable the model to aggregate verse-level predictions into a single attribution at the poem level, similar to how scholars would analyze poems as coherent artistic wholes made of interdependent beyts.

Finally, a verse-level granularity is in keeping with current practices in literary research, in which scholars often examine single, discrete beyts to explicate their thematic, rhetorical, or symbolic significance. This fine level of granularity will enable our system to capture the complex dynamics of literary analysis and generate outputs that are both interpretable and meaningful for humanistic inquiry.

\subsubsection{Data Splitting Strategy}
\label{Data Splitting Strategy}
A major concern in the creation of our poet classification model is data leakage mitigation, specifically since our model works at the level of the verse (\textit{beyt}), where a single verse is presented as an isolated input sample that carries a unique unit of sense, rhythm, and style. However, verses in a poem often share thematic material, stylistic features, and sometimes even the same vocabulary or metrical structure, leading to the likelihood that verses from a single poem will be in both the test and train set, potentially creating an artificial boost in the performance metric. Data leakage is a phenomenon whereby the model ends up retaining knowledge from partially seen poems in the training phase and applying this learned knowledge incorrectly to predict verses from the same poem during the test phase.

\vspace{0.5em}
\noindent
\textbf{Poem-Level Partitioning.}
In addressing this challenge, we undertook a strategy of poem-level splitting, which ensures lines in a particular poem are only placed in either the train, validation, or test set. It keeps the model from ever experiencing any poems in the train stage that will be tested in later stages. This is in keeping with the intrinsic nature of Persian poetry in that beyts in a poem often carry relationships through imagery, metaphorical frameworks, and thematic coherence. Our evaluation is therefore an effective measure of the model's true ability to generalize to unseen poems.

\vspace{0.5em}
\noindent
\textbf{Stratified Sampling by Poet Label.}
To ensure fair representation of the different classes in the different segments of the dataset, and to avoid the learning process having a predominance towards more frequently occurring poets, stratified sampling according to poet labels at the poem level was applied. This approach ensures each poet's representation is seen in the training, evaluation, and test datasets so that model performance can be fairly and meaningfully compared across all the poets regardless of their comparative prevalence in the collection.

Divisibility Pairs
The data set was divided into:

Training Set (80\%): Used to identify the distinctive features that separate one poet from another by examining the textual, semantic, and stylistic features of a particular line.

Validation Set (10\%): Used during hyperparameter optimization and to monitor overfitting during the training process to ensure model robustness.

Test Set (10\%): Completely held out during training and validation, used for final performance evaluation.

These proportions strike a balance between providing sufficient data for learning while reserving enough samples for unbiased evaluation.

Class Imbalance Considerations.
Despite the stratification, Persian poetry datasets naturally exhibit a class imbalance, with some poets contributing a large number of verses and others contributing only a small number. To ensure fair representation of all poets and to avoid the possibility of unreliable and misleading assessment, we have removed poets whose contributions comprise less than 50 verses from the dataset. The threshold provides a minimum sample size that allows the model to properly learn and evaluate style features peculiar to each poet.

Aggregation techniques Given that the model makes a prediction per verse, it was critical to develop a systematic and understandable strategy to aggregate verse-level predictions to make decisions on the poem level. Majority voting, where the poem-level prediction is made based on the most dominant poet label of the verses, was coupled with probability-based voting. This latter strategy combines the model's confidence scores with respect to the predicted labels of particular verses to make a global decision. These methods leverage the fine grain in our verse-level data while staying faithful to the poem level structure, a salient aspect of Persian poetry.

\vspace{0.5em}
\noindent

\subsection{Problem Formulation}

The major aim of the current research is the automated classification of Persian poetry based on its respective authors, presented as a multi-class classification problem at the level of verses. Due to the historical significance of Persian poetry, in addition to the various stylistic, semantical, and metrical features that characterise an individual poet's work, this project is not merely a computational task but contributes uniquely to the broader field of literary analysis, in which queries regarding authorship authentication and analysis of stylistic features are of paramount significance.

\vspace{0.5em}
\noindent
\textbf{Input and Output Definitions.}
Let $\mathcal{P} = {P_1, P_2, \dots, P_N}$ be the set of the poets we are analyzing, where $N$ is the number of the participating poets. Every poem in our collection is composed of a single or several verses (\textit{beyts}), with one $b_i$ verse being treated as an independent data example. In addition, a verse is defined by a set of feature modalities:
\begin{itemize}
\item \textbf{Tokenized Text:} A sequence of subword tokens representing the content of the verse, following a maximum of 64 tokens to well capture both semantic and syntactic features.
\item \textbf{Semantic Embeddings:} A 100-dimensional vector that captures distributional semantics from Word2Vec embeddings trained on the given corpus.
\item \textbf{Stylometric Features:} A 7-dimensional vector encoding quantitative measures of writing style, including word count, lexical diversity, punctuation density, and symmetry ratio.
\item \textbf{Structure and Meter:} One-hot encoded vectors representing the poetic structure and meter classification, respectively.
\end{itemize}

Using the mentioned attributes, the aim is to define the function $f: b_i \rightarrow P_j$ that matches a verse $b_i$ with its most likely poet $P_j \in \mathcal{P}$.

\vspace{0.5em}
\noindent
\textbf{Verse-Level Classification.}
We use a classification system at the verse level to take advantage of the unique structure found in Persian poetry, where each \textit{beyt} usually has its own thematic, rhetorical, and stylistic significance. This allows close analysis of authorship and aligns with literary conventions that tend to analyze beyts separately.

However, modeling at the verse level also presents challenges:
\begin{itemize}
\item Verses are often brief, giving little context in distinguishing a poet from another.
\item Poets have the capacity to change their stylistic mode, structural framework, and rhythmic modes either in one poem or in several poems.
\item Shared Literary Traditions: Some poets intentionally mimic the style of others, making authorship attribution challenging.
\end{itemize}

\vspace{0.5em}
\noindent
\textbf{Poem-Level Aggregation.}
Though the model is per-verse trained and tested, it is still possible to aggregate predictions at the level of the entire poem, thus improving interpretability and enabling practical evaluation. Because all verses in a poem are assigned the same data split, this can be achieved through voting schemes like majority voting or probability-based methods. This approach recognizes the coherence in poem composition and guards against data leakage, thus allowing the model's evaluation to truly capture its ability to generalize to unseen, novel poems.

\vspace{0.5em}
\noindent
\textbf{Ambiguity and Controversial Verse.}
A unique challenge in the domain of Persian literary studies involves the presence of disputed poems with different attributions across manuscripts. Though such cases are excluded from training processes to ensure the validity of ground-truth labels, they are included for later evaluation tasks. This setup allows for an investigation into the model's ability to resolve unclear authorship, a task that is of great importance in the digital humanities field.

\vspace{0.5em}
\noindent
\textbf{Objective Function.}  
The classification model is trained using a standard cross-entropy loss, where the ground-truth poet label is encoded as a one-hot vector and compared against the model’s predicted probabilities. Formally, let $y_i$ denote the true poet label for verse $b_i$, and $\hat{y}_i$ the predicted probability distribution over poets. The objective is:
\[
\mathcal{L} = -\sum_{i=1}^{N} y_i \log(\hat{y}_i).
\]

\vspace{0.5em}
\noindent

\subsection{Model Architecture}

\subsubsection{Overview}

The suggested architectural model is intended to classify Persian poetry by author at the verse (\textit{beyt}) level. Given the complex stylistic and structural elements that are inherent in Persian poetry, the model leverages state-of-the-art contextual language representations in combination with carefully developed literary features to achieve a high degree of accuracy in classification. Essentially, the model uses a transformer-based text encoder for extracting contextual embeddings from the verse text, thus capturing the semantic and syntactic nuances that reflect the unique writing style of a poet \cite{vaswani2017attention,salami2021recurrent}.

In addition to the text encoder, the model incorporates multiple attributes that cover a broad range of stylistic paradigms, such as semantic embeddings, stylometric features, poetical setups, and meter schemes \cite{rezaei2017stylometric,hamidi2011meter,salami2021recurrent}. Combining the different attributes allows the model to fully describe a single verse, thus enriching localized context with global stylistic markers. The multitask structure allows the model to observe fine-grained distinctions between authors who might utilize similar vocabularies but have different preferences in terms of meter or stylistic preferences.

\subsubsection{Text Encoder}

The text encoder is our model's backbone and is the central component in how the semantic, syntactic, and rhetorical complexity typical of Persian poems is well captured. It takes in raw tokenized lines and transforms them to context-aware embeddings that capture local and global word-to-word relations, thus enabling the downstream classifier to differentiate authors according to their characteristic writing styles \cite{vaswani2017attention}.

\vspace{0.5em}
\noindent
\textbf{Drivers and Barriers.}
Classical Persian poems are characterized by their adoption of complex rhetorical devices, metaphorical terms, and complex syntactic constructions. Unlike a typical narrative structure of a piece of prose, the poem often uses an elaborate network of complex metaphors, personifications, and multi-faceted meanings communicated through a string of many words or lines. Such natural complexity poses significant challenges to any machine learning model in understanding the poem's meaning and linking it to a specific poet. To mitigate those challenges, our study uses an encoding approach rooted in the transformer model, generally recognized to be able to model long-distance dependencies and make use of bidirectional attention, which are most beneficial in revealing complex relationships in textual data \cite{vaswani2017attention}.

\begin{figure}[ht]
\centering
\includegraphics[width=0.5\linewidth]{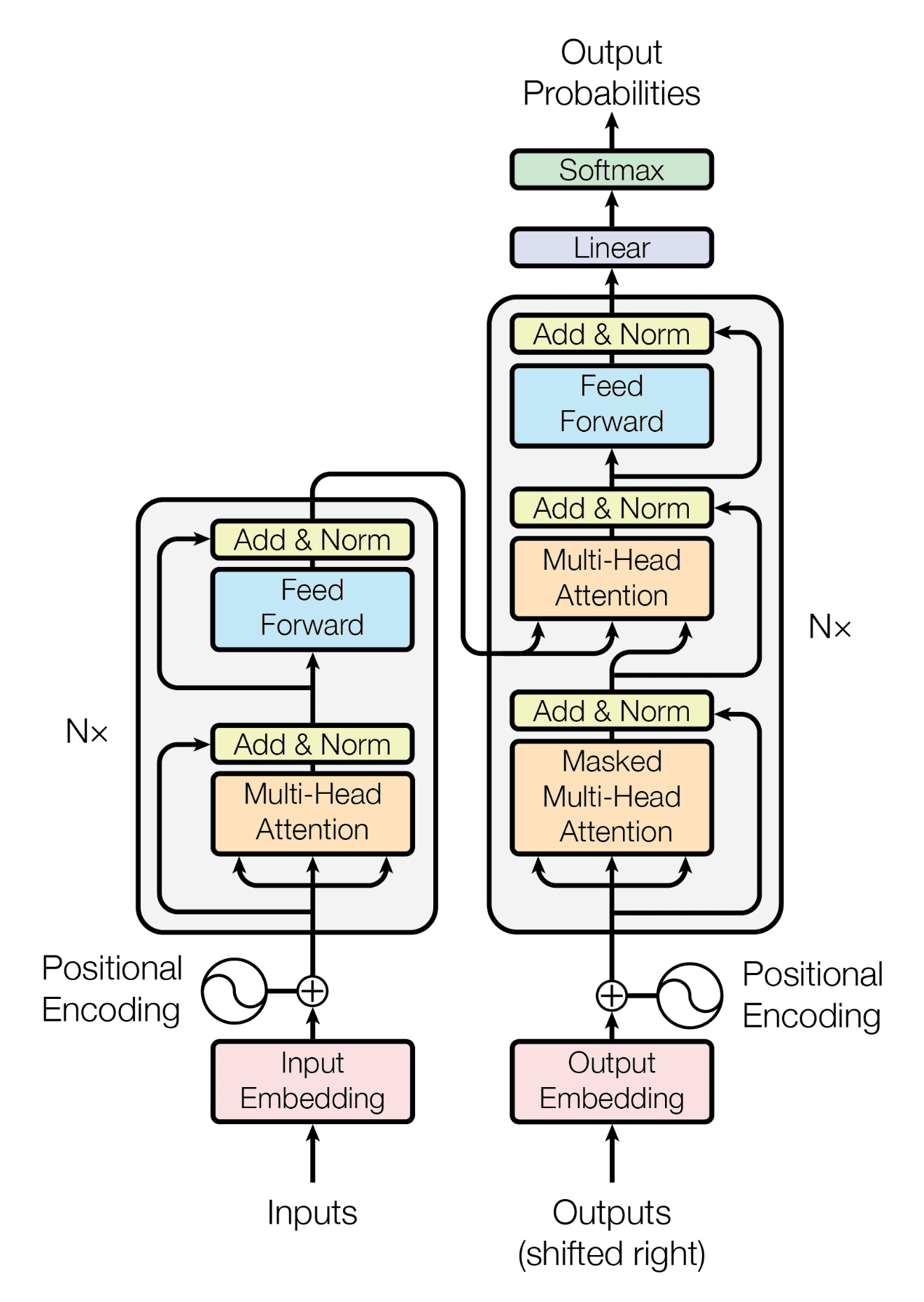}
\caption{The encoder-decoder structure of the Transformer architecture.}
\label{fig:meter-distribution}
\end{figure}

\vspace{0.5em}
\noindent
\textbf{Transformer Encoder Fundamentals.}
The transformer encoder, initially presented in Vaswani et al. is based on the self-attention scheme, allowing all tokens in the input sequence to attend to all the other tokens simultaneously\cite{vaswani2017attention}. This is a significant departure from traditional RNN or CNN models, where a sequential process or local receptive fields respectively impose strictures. In the transformer encoder, every layer is made up of two major sub-layers:

\begin{itemize}
\item \textbf{Multi-Head Self-Attention.}
The multi-head self-attention mechanism enables the model to discover different types of relationships between tokens. Each attention head computes a scaled dot-product attention:
    \[
    \text{Attention}(Q, K, V) = \text{softmax}\left(\frac{QK^\top}{\sqrt{d_k}}\right)V
    \]
    Here, $Q$, $K$, and $V$ refer to the query, key, and value matrices, respectively, obtained from the input token embeddings. Also, $d_k$ represents the key vector dimension. Involving multiple attention heads allows the model to focus simultaneously at different locations and pay attention to different linguistic features, such as syntactic structure, semantic alignments, and rhetorical features.

\item \textbf{Position-Wise Feed-Forward Network.}
Following the self-attention mechanism, each token's representation is processed by a fully connected feed-forward network, which is applied separately at each position:
    \[
    \text{FFN}(x) = \text{ReLU}(xW_1 + b_1)W_2 + b_2
    \]
    where $W_1$, $W_2$, $b_1$, and $b_2$ are learned parameters. This sub-layer allows the model to capture complex feature interactions and enhances its expressive capacity.
\end{itemize}

Every sub-layer in the transformer encoder includes residual connections and layer normalization, enabling stable training and gradient flow. Through the stacking of multiple layers, the representational capacity of the model is increased, enabling the learning of hierarchical features from text data.

\vspace{0.5em}
\noindent
\textbf{Positional Encoding.}
Without an intrinsic knowledge of token sequences, input embeddings are complemented with positional encodings in order to encode information about the relative and absolute position of tokens in a sequence. These encodings can be static (e.g., based on sinusoidal functions) or learned during the process of training. This is specifically applicable in the case of Persian poetry, where the influence of surrounding words and sentences often is determinant, and thus the inclusion of positional information is essential in maintaining the structure and rhythm of the verses.

\vspace{0.5em}
\noindent
\textbf{Bidirectional Attention.}
One major feature of the transformer encoder is its bidirectional self-attention structure, through which a single token can see all tokens before and after it in a sequence \cite{vaswani2017attention}. This is in sharp contrast to traditional unidirectional models based on sequential processing of tokens and the use of contextual information from a single direction only. In the case of Persian poetry, having access to context in both directions is especially crucial, since meanings in this genre often arise from interactions between words, hemistiches, and literary devices spread across the poem. As an example, the second hemistich of a \textit{beyt} can clarify or boost the significance of the first hemistich, so that a model must have a holistic understanding of contextual connections in both directions.

\vspace{0.5em}
\noindent
\textbf{Aggregation and Presentation of Results.}
Following the generation of context-aware embeddings of a target token by the transformer encoder, a pooling technique is applied in order to obtain a fixed-dimensional representation of the verse. The most common practice is to use the hidden state of the first token in the input string, often a [CLS] token or an equivalent. This vector is a summary embedding, capturing details of the entire verse together with localized lexical features and the global poem context. This representation is then passed to the classifier head along with auxiliary features in order to enable the final author prediction.

\vspace{0.5em}
\noindent
\textbf{The Suitability of Transformer Encoders for Persian Poetry Analysis.}
Transformer encoders exhibit improved results in the task of classifying Persian poetry due to a range of different factors:

\begin{itemize}
\item \textbf{Extended Dependencies.} Persian poems often illustrate dependencies that span many tokens, represented by metaphors that appear over several words or by literary devices that connect parts of a poem. Self-attention makes it possible for the model to capture such long-distance interactions \cite{vaswani2017attention}.
\item \textbf{Bidirectional Context.} Poetic language often becomes meaningful through the interactions between different parts of a poem or a stanza. Bidirectional attention allows the model to handle both previous and future dependencies, both of which are essential to understanding the context of a poem.
\item \textbf{Parallel Processing.} Transformers can process all tokens in a sequence simultaneously, enabling faster training compared to sequential models like RNNs.
\item \textbf{Flexibility.} Transformers are extremely flexible, enabling the incorporation of ancillary features, and are versatile in being applied to other modalities, such as metrical and stylometric information, required in the analysis of Persian poems.
\end{itemize}

\vspace{0.5em}
\noindent
\textbf{Relationship to BERT and RoBERTa.}
The development of our text encoder is largely based on transformer-based models, in particular BERT and RoBERTa, that are pre-trained on massive masked language modeling tasks before being fine-tuned on downstream tasks \cite{vaswani2017attention,alqurashi2025bert,qarah2024arapoembert}. Although both models share the same transformer encoder, the pre-training goals and underlying datasets are different. BERT includes a next sentence prediction task, generally considered less beneficial in many cases; in contrast, RoBERTa removes this task, uses dynamic masking, uses larger batches, and utilizes a larger pre-training corpus to optimize its effectiveness. In the context of Persian poetry, the models provide a sturdy starting point, providing dense contextual representations that well capture both the literal and metaphorical meanings of poetic language.

\subsubsection{Auxiliary Feature Inputs}

While the transformer-based text encoder successfully produces rich contextual representations of the poem's tokens, it is also vital to complement this with extra features capturing different aspects of style, structure, and semantics—something that is of particular importance when dealing with Persian poetry, where authors will often develop similar themes while at the same time expressing differentiating nuances in writing style, meter, and form. Such features act as ancillary signals that enhance the model's ability to differentiate between authors, in particular in complex situations where the textual information in isolation might be insufficient.

\vspace{0.5em}
\noindent
\textbf{Justification for Auxiliary Characteristics.}
Persian poetry is a rich literary tradition in which understanding goes beyond mere analysis of isolated words and expressions; it is strongly conditioned by formal elements like meter (\textit{aruz}), the general structure of the poem, and stylometric features like diction and punctuation. Therefore, text embeddings in isolation might lead to the exclusion of crucial stylometric markers that lie at the root of truly understanding authorship. In an attempt to mitigate this limitation, we examine four types of auxiliary features: semantic embeddings, stylometric features, poem structure elements, and meter patterns. Each of these types captures a distinct but complementary aspect of the poem, and thus contributes to the characteristic literary signature of a particular poet.

\vspace{0.5em}
\noindent
\textbf{Semantic Representations.}
Although the transformer-based encoder exhibits a capacity to recognize complex contextual relations between tokens, its focus is still largely on the local context of isolated verses. To extend this, we add a 100-dimensional semantic embedding vector, gained from word embeddings learned through Word2Vec over a vast collection of Persian poems. Word embeddings are based on the hypothesis of distributions, where words occurring in similar contexts are likely to share similar meanings; this captures broader semantic relations at the level of the entire corpus, transcending the limits of single poems. This allows the model to leverage distributional semantics based on lexical associations, thematic clusters, and co-occurrence patterns that often are used by different types of poets. Some poets, for example, will prefer words with mystical or romantic themes, while others might prefer philosophical or instructional themes. Adding in these semantic embeddings allows the model to pick up on subtler usage patterns that may not be visible from a single line of poetry.

\vspace{0.5em}
\noindent
\vspace{0.5em}
\noindent
\textbf{Stylometric Features.}
Stylometry, or the quantitative analysis of writing style, is a method that has been applied over a long period of time in the field of authorship attribution, stemming from initial research in computational linguistics. In Persian poetry analysis, stylometric features are of particular importance because of their stable and quantifiable stylistic patterns that persist throughout the work of a poet. For every verse, we extract seven fundamental stylometric features, each carefully designed to capture an independent aspect of authorship style:

\begin{itemize}
\item \textbf{Word Count:} Total number of words in the verse that reflects either verbosity or brevity.
\item \textbf{Distinct Word Count:} The count of unique words, indicating both lexical diversity and creative expression.
\item \textbf{Average Lexical Length:} A measure of both linguistic selection and morphological complexity.
\item \textbf{Hapax Legomena Ratio:} A proportion of words that appear only once in the verse, reflecting both the freshness and variety of vocabulary.
\item \textbf{Mean Hemistich Length:} The length of an average hemistich (half-verse) is an indication of rhythmic balance and symmetry, both being necessary elements of Persian meter.
\item \textbf{Punctuation Density:} Punctuation mark to verse length ratio, reflecting the organizational rhythm and pacing in the rhetoric.
\item \textbf{Symmetry Ratio:} The comparative lengths of the two hemistiches, capturing the balance in structure of the verse—a necessary part of the Persian poem's aesthetic structure.
\end{itemize}

Every stylometric feature is normalized to ensure consistency with different poets and to reduce the impact of attributes that have a wider range of values on the learning process. In addition, normalization helps in stabilizing the learning process and speeds up convergence.

\vspace{0.5em}
\noindent
\textbf{Stanza Construction.}
The formal structure of poetic genres is a defining feature of Persian poetry, having a profound impact on thematic material as well as rhythm, rhyme patterns, and the poetological devices employed by the poet. Our anthology represents a wide range of poetic genres, such as but not limited to \textit{ghazal}, \textit{masnavi}, \textit{qasida}, \textit{rubāʿī}, and \textit{do-beyti}. Each of these genres has its own traditions regarding structure, rhymes, and thematic focus. \textit{Ghazals}, for example, often explore themes of mysticism and love, while \textit{masnavis} are typically marked by the telling of epic or didactic tales.

To represent this structural aspect, the form of each poem is represented in a one-hot vector. Every member of this vector represents a unique form found in the data set. Using this approach allows the model to differentiate verses based on different forms and to draw upon knowledge that an author will be good in a particular form or will shift based on the form taken. Through the inclusion of the form, the model gains knowledge of the structural conventions that govern the works of any poet.

\vspace{0.5em}
\noindent
\vspace{0.5em}
\noindent
\textbf{Rhythmic Organization.}
Meter, or \textit{aruz}, is a central component of Persian poetry, determining the rhythmic pattern, musicality, and overall aesthetic merit of poetic lines. Each meter has a unique pattern of syllables and stresses, and poets often show preferences for certain meters or use them to achieve certain emotional effects. Our dataset contains \textbf{123} individual metrical patterns, which we reduce to \textbf{15 standardized meter classes} to achieve a balance between granularity and computational tractability.

The meter in any poem is represented via a one-hot encoding strategy, allowing the model to identify rhythmic configurations that can serve as a style signature unique to a range of different poets. This feature is of great value, since authors typically choose a particular meter to suit their thematic or emotional goals; hence, meter can provide vital information about authorship. A poet might employ a particular meter in \textit{ghazals} and a different one in \textit{masnavis} or \textit{qasidas}, for example. In representing metrical features as a separate feature, we increase the model's ability to identify such authorial preferences and combine them with text and semantic features.

\vspace{0.5em}
\noindent
\textbf{Feature Integration Strategy.}
Each auxiliary feature vector, consisting of semantic, stylometric, formal, and metrical features, is processed in isolation before being integrated into the transformer-based text representation. This design ensures that the model's representation of the entire verse is complemented by every feature category while retaining the unique information contained in each modality. Concatenation of the next layer yields a high-dimensional feature vector, which is further fed into the classifier head. Multi-input design allows the blend of localized contextual information from the text encoder with global stylometric and structural features to form a more expressive and discriminative representation of the entire verse.

\vspace{0.5em}
\noindent
\textbf{Multi-Modal Integration.}
The Multi-Input Fusion module is the central point where different representations of one verse—obtained from the transformer-based text encoder and auxiliary feature modules—are fused into a unified vector. This fusion is essential in addressing the complex nature of Persian poetry, where sense, style, and form are the results of an intricate interplay of lexical, semantic, rhythmic, and structural elements. In this subsection, we provide a thorough and detailed analysis of the multi-input fusion process, including the underlying rationale, implementation details, and the linkage to the task of classification.

\vspace{0.5em}
\noindent

\textbf{Reasons and Motivation.}
Persian poetry goes beyond linguistic analysis since it is a carefully structured literary form that is based on multiple aspects of significance and expression. Every verse (\textit{beyt}) inherently contains:

\begin{itemize}
\item \textbf{Lexical Semantics:} The careful choice of words, phrases, and rhetorical devices.
\item \textbf{Contextual Semantics:} The word relationships established by the particular context of the poem, including syntactic relations and the use of figurative language.
\item \textbf{Stylistic Features:} The consistent writing habits of poets, such as word count, lexical diversity, and punctuation usage.
\item \textbf{Structural Features:} Poem structure and meter, both of which together determine the poem's rhythm, aesthetic characteristics, and thematic conventions.
\end{itemize}

While the transformer encoder is seen to perform well in understanding both contextual and local semantics in separate verses, it can struggle to fully understand the complex stylistic and compositional features required to identify different poets. Combining such features with the transformer output allows the model to utilize further information to increase its discriminatory power.

\vspace{0.5em}
\noindent
\textbf{Concatenation Methodology.}
The fusion process begins with the combination of the feature vectors produced by each module:

\begin{itemize}
\item The transformer-based text encoder produces a 768-dimensional contextual embedding vector, $\mathbf{h}{\text{text}}$, representing the semantic meaning of the verse.
\item The semantic embedding module produces a 100-dimensional vector, named $\mathbf{h}{\text{semantic}}$, representing the overall distributional semantics.
\item The stylometric features produce a 7-dimensional vector, $\mathbf{h}{\text{stylometric}}$, representing stylistic inclinations.
\item The form and meter components produce one-hot encoded vectors, referred to as $\mathbf{h}{\text{form}}$ and $\mathbf{h}_{\text{meter}}$, respectively, representing the structural genre and rhythmical pattern of the poem.
\end{itemize}

The vectors are concatenated along the feature dimension to obtain a unified, high-dimensional representation:
\[
\mathbf{h}_{\text{concat}} = [\mathbf{h}_{\text{text}} \, \Vert \, \mathbf{h}_{\text{semantic}} \, \Vert \, \mathbf{h}_{\text{stylometric}} \, \Vert \, \mathbf{h}_{\text{form}} \, \Vert \, \mathbf{h}_{\text{meter}}].
\]
This framework maintains the unique information contained in each modality, but simultaneously allows the model to grasp their interactions in the deeper layers.

\vspace{0.5em}
\noindent
\textbf{Advantages of Feature Integration.}
The combination of multiple input modalities into a single representation offers the model many advantages:

\begin{itemize}
\item \textbf{Complementarity:} This term refers to the idea that different types of features capture different aspects of poetic style. For example, meter captures rhythmic and structural elements, while thematic relations are conveyed semantically through embeddings.
\item \textbf{Increased Capacity for Distinction:} While some poets may use similar linguistic mechanisms, they often differ in their use of meter or stylistic techniques. Incorporating such features enables the model to distinguish between poets who might otherwise appear similar based solely on textual content.
\item \textbf{Usability:} Concatenation is a straightforward process and has the flexibility to incorporate new types of features as future research identifies additional literary elements.
\end{itemize}

\vspace{0.5em}
\noindent
\textbf{Alternative Fusion Methodologies.}
Although concatenation is the most obvious solution, we have also considered other methods, including:

\begin{itemize}
\item \textbf{Attention-based Fusion:} Uses attention mechanisms to evaluate the relative importance of different modalities in relation to the input verse in a dynamic context.
\item \textbf{Gated Fusion:} Uses adaptive gates to control the flow of information from each feature module, similar to the gating techniques used in LSTM networks.
\item \textbf{Cross-Modality Interaction:} Employs bilinear pooling or tensor fusion methods to explicitly model interactions between different modalities.
\end{itemize}

While these methods hold the potential for future benefits, they also introduce additional computational complexity and require careful tuning to avoid overfitting, particularly in the context of high-dimensional text data and ancillary features.

\vspace{0.5em}
\noindent
\textbf{Implementation Considerations.}
From our empirical analysis, we found that using simple concatenation with a fully connected classification head achieves an optimal balance between model performance and complexity. This architecture allows the model to learn the relative importance of different features at various stages of training, enabling it to focus on specific features relevant to both the poet and the verse. The concatenated vector is then passed through a feed-forward neural network with non-linear activation functions (e.g., ReLU), enabling the model to capture complex feature interactions and higher-order relationships relevant to authorship attribution.

\vspace{0.5em}
\noindent
\textbf{The Importance of Fusion in Persian Poetic Tradition.}
Persian poetry is a rich combination of semantics, rhythm, and stylistic features. A poem written by Hafez, compared to one written by Rumi, can utilize the same vocabulary but share incredibly different metrics, structure, or fine stylistic choices. In the event of no fusion, the model would be ignoring these subtleties and concentrating too much on superficial lexical features. Incorporating auxiliary features into the model introduces the capability to identify both obvious and subtler stylistic differences, thus increasing the model's power to provide correct and reliable predictions about authorship.

\vspace{0.5em}
\noindent
\textbf{Classification Tool.}
The Classification Head is the final and central element of our architectural design, enabling the mapping of complex and heterogeneous representations of a verse to a probability distribution over multiple genres of poems. This module allows the transformation of high-dimensional feature vectors into outputs that are amenable to interpretation, while it also contains key decisions affecting the model's learning dynamics, expressiveness, and generalizability. In the following, we provide an in-depth discussion about the Classification Head, including its compositional structure, functionality, and rationale behind decisions made in the design.

\vspace{0.5em}
\noindent
\textbf{Objectives and Purposes.}
The sum of transformer-based contextual embeddings and auxiliary features such as semantic embeddings, stylometric features, poem form, and meter extracted from the Multi-Input Fusion module leads to creation of a merged feature vector referred to as $\mathbf{h}_{\text{concat}}$. It combines both rich contextual details and wider stylistic features. The role of the Classification Head is to process this vector and produce a probability distribution over the candidate poet classes. Algebraically, based on the unified representation:

\[
\mathbf{h}_{\text{concat}} \in \mathbb{R}^{d}
\]
where $d$ is the total dimensionality of the concatenated features, the Classification Head computes:
\[
\hat{\mathbf{y}} = \text{Softmax}(f(\mathbf{h}_{\text{concat}}))
\]
where $f(\cdot)$ is a learnable function parameterized by neural network layers.

\vspace{0.5em}
\noindent
\textbf{Layered Structure.}
To achieve the transformation from $\mathbf{h}_{\text{concat}}$ to $\hat{\mathbf{y}}$, we design the Classification Head as a multi-layer feed-forward neural network. This layered architecture is motivated by the need to:
\begin{itemize}
    \item Capture non-linear interactions between features that might not be linearly separable.
    \item Model complex relationships between textual semantics and auxiliary features, such as interactions between metrical preferences and stylistic habits.
    \item Provide sufficient capacity to learn discriminative patterns that separate different poets.
\end{itemize}

The architecture comprises the following components:

\begin{itemize}
    \item \textbf{Fully Connected Linear Layer.}
    The first layer projects the high-dimensional input vector into a lower-dimensional space, typically reducing the feature dimension from $d$ to an intermediate size (e.g., 512 units). This dimensionality reduction serves two purposes: (a) it enables the model to learn a compact representation that captures essential interactions, and (b) it reduces computational complexity in subsequent layers. Mathematically:
    \[
    \mathbf{h}_1 = \text{ReLU}(\mathbf{W}_1 \mathbf{h}_{\text{concat}} + \mathbf{b}_1)
    \]
    where $\mathbf{W}_1 \in \mathbb{R}^{512 \times d}$ and $\mathbf{b}_1 \in \mathbb{R}^{512}$ are learnable parameters.

    \item \textbf{Non-Linear Activation.}
    The ReLU (Rectified Linear Unit) activation function introduces non-linearity into the model, allowing it to capture complex feature interactions that are essential for distinguishing between poets with subtle stylistic differences. ReLU is defined as:
    \[
    \text{ReLU}(x) = \max(0, x)
    \]
    and is widely used due to its simplicity, efficiency, and effectiveness in mitigating the vanishing gradient problem.

    \item \textbf{Dropout Regularization.}
    To prevent overfitting—especially important given the high-dimensional input and the relatively small number of training examples per poet—we apply dropout regularization after the activation function. Dropout randomly zeroes a fraction $p$ of the elements in the input vector during training, forcing the model to learn redundant representations and improving its generalization capability. Typically, we set $p = 0.3$:
    \[
    \mathbf{h}_1^{\text{drop}} = \text{Dropout}(\mathbf{h}_1, p=0.3)
    \]

    \item \textbf{Output Layer.}
    The final layer is a linear transformation that projects the intermediate representation to a vector of logits, one per poet class:
    \[
    \mathbf{z} = \mathbf{W}_2 \mathbf{h}_1^{\text{drop}} + \mathbf{b}_2
    \]
    where $\mathbf{W}_2 \in \mathbb{R}^{C \times 512}$, $\mathbf{b}_2 \in \mathbb{R}^{C}$, and $C$ is the number of poet classes. The output logits are then passed through a softmax function to produce a valid probability distribution over the classes:
    \[
    \hat{\mathbf{y}}_i = \frac{\exp(z_i)}{\sum_{j=1}^{C} \exp(z_j)}
    \]

    This output vector $\hat{\mathbf{y}}$ represents the model’s confidence in assigning the verse to each poet.

\end{itemize}

\vspace{0.5em}
\noindent
\textbf{Training Objective.}
The model is trained using the cross-entropy loss function, which measures the dissimilarity between the predicted probability distribution $\hat{\mathbf{y}}$ and the ground-truth one-hot label vector $\mathbf{y}$:
\[
\mathcal{L} = -\sum_{i=1}^{C} y_i \log(\hat{y}_i)
\]
where $y_i \in \{0, 1\}$ and $\sum_{i} y_i = 1$. This loss encourages the model to assign high probability to the correct poet class while minimizing the probability assigned to incorrect classes.

\vspace{0.5em}
\noindent
\vspace{0.5em}
\noindent
\textbf{Interpretability and Assurance Metrics.}
One major advantage of having a softmax output layer is the interpretability of its outputs since the output of the model can be interpreted as confidence scores. The confidence scores can then be combined at the poem level under different voting strategies, including majority voting or probability voting, to enable trustworthy authorship predictions. In addition, the method enables the application of thresholding to recognize predictions below a certain confidence threshold as uncertain, which is a critical feature in the case of disputed or unclear poems.

\vspace{0.5em}
\noindent
\textbf{Design Rationale.}
The Classification Head is designed to achieve a balance of complexity and interpretability. Through the utilization of a simple but effective fully connected structure, both textual and auxiliary features may be exploited without imposing excessive computational costs. The use of dropout and ReLU activation further enhances generalizability and stability during training, while the softmax output provides interpretable and understandable probability estimations. In addition, modularity allows easy ablation study execution, enabling systematic assessment of the contributions of the components of the input features towards the model's final performance.

\vspace{0.5em}
\noindent
\textbf{Training Objective.}
The training target acts as the model's educational backbone, guiding model parameter optimization towards maximizing accuracy in distinguishing the authorship of poetical works in Persian. In this section, a thorough and systematic explanation of the training target is provided, including the theoretical bases, implementation details, and design aspects that align with the specific challenges of modeling classical Persian poetry.

\vspace{0.5em}
\noindent
\textbf{Objective of Training.}
The main purpose of the training process is to tune the model's tunable parameters, such as the weights of the transformer encoder, different feature processing components, and the classification head. This tuning is intended to allow the model to correctly map an input verse—its textual, semantic, stylometric, formal, and metrical features—onto the corresponding poet label. Mathematically, the goal is to find a function $f_\theta(\cdot)$, parameterized by $\theta$, which optimizes the reduction in the difference between the predicted probability distribution over poets $\hat{\mathbf{y}}$ and the true ground-truth label $\mathbf{y}$.

\vspace{0.5em}
\noindent
\vspace{0.5em}
\noindent
\textbf{Cross-Entropy Loss.}
Given the multi-class nature of the problem in classification—each verse being classified into one of the $C$ classes of poets—we utilize the categorical cross-entropy loss function. This particular loss function measures the difference between the true distribution (expressed in a one-hot encoded form) and the estimated distribution (given by the softmax function in the classification head). Mathematically, in the case of a particular example in the training set:

\[
\mathcal{L}_{\text{CE}}(\hat{\mathbf{y}}, \mathbf{y}) = -\sum_{i=1}^{C} y_i \log(\hat{y}_i)
\]
where:
\begin{itemize}
    \item $\mathbf{y} = [y_1, y_2, \ldots, y_C]$ is the one-hot encoded ground-truth label vector, with $y_i = 1$ for the correct class and $0$ otherwise.
    \item $\hat{\mathbf{y}} = [\hat{y}_1, \hat{y}_2, \ldots, \hat{y}_C]$ is the predicted probability vector output by the model.
\end{itemize}

The cross-entropy loss penalizes the model for assigning low probability to the correct class and high probability to incorrect classes. Minimizing this loss encourages the model to produce confident and accurate predictions.

\vspace{0.5em}
\noindent
\textbf{Batch-Level Loss Computation.}
Given that the model is trained using mini-batch gradient descent, the total loss for a batch of size $N$ is computed as the mean of individual losses:
\[
\mathcal{L}_{\text{batch}} = \frac{1}{N} \sum_{j=1}^{N} \mathcal{L}_{\text{CE}}(\hat{\mathbf{y}}^{(j)}, \mathbf{y}^{(j)})
\]
where $\hat{\mathbf{y}}^{(j)}$ and $\mathbf{y}^{(j)}$ denote the predicted and ground-truth label vectors for the $j$-th sample in the batch. This averaging ensures that the loss scale remains consistent regardless of batch size and facilitates stable training.

\vspace{0.5em}
\noindent
\textbf{Incorporating Class Imbalance.}
Persian poetry datasets often exhibit class imbalance, with some poets contributing thousands of verses and others only a few hundred. This imbalance can bias the model towards majority classes, leading to poor performance on underrepresented poets. To mitigate this, we incorporate class weights into the loss function:
\[
\mathcal{L}_{\text{weighted}} = -\sum_{i=1}^{C} w_i y_i \log(\hat{y}_i)
\]
where $w_i$ is the weight assigned to class $i$, typically set as the inverse of the class frequency in the training set. This reweighting ensures that the model pays proportionate attention to all classes, fostering fair and balanced learning across poets.

\vspace{0.5em}
\noindent
\textbf{Regularization Techniques.}
To counter the danger of overfitting, especially with the small number of verses available for some poets, we introduce regularization methods that augment the cross-entropy objective:
\begin{itemize}
    \item \textbf{Dropout:} Applied in the middle layers of the classification head, dropout randomly eliminates a proportion of input units in the process of model training, forcing the model to learn robust and overlapping representations that generalize well to previously unseen data.
    \item \textbf{Weight Decay:} Implemented via $L_2$ regularization, weight decay adds a penalty to high-magnitude weights, encouraging smoother and less drastic updates of the model parameters.
\end{itemize}
These methods complement the main loss function, enabling better generalization and improving model resilience.

\vspace{0.5em}
\noindent
\textbf{Optimization Algorithm.}
The overall training objective is minimized using a variant of stochastic gradient descent, commonly referred to as the Adam optimizer, which combines momentum with adaptive learning rates. At each iteration, the optimizer updates the model parameters $\theta$ by computing the gradients of the loss function with respect to these parameters:

\[
\theta \leftarrow \theta - \eta \cdot \nabla_\theta \mathcal{L}
\]
where $\eta$ is the learning rate. Adam’s adaptive learning rate mechanism accelerates convergence by scaling the learning rate based on past gradients.

\vspace{0.5em}
\noindent
\textbf{Proactive Termination and Model Storage Evaluation.}
Given the propensity of overfitting, particularly in classification tasks with high dimensionality, we use early stopping based on validation accuracy. Training is stopped when no improvement in the validation accuracy is observed after a specified number of epochs (known as patience), and at that time the model having the best accuracy on the validation set is retained. This ensures that the model is optimally functional on unseen data and not just on the noise contained in the training set.

\vspace{0.5em}
\noindent
\textbf{Abstention and Confidence Thresholding.}
In real-life examples of literary analysis, especially when faced with authorship disputes that may be unclear or contentious, it is often wise for the model not to make a prediction when the degree of confidence is too low. Thresholding can be applied simply in practice using the softmax function; that is, if the highest probability prediction is less than a predetermined threshold value $\tau$, the model is best placed not to make a prediction. This makes it extremely helpful in follow-on evaluation jobs and is aligned with humanistic research paradigms, where ambiguity is preferred to strict classifications.

\vspace{0.5em}
\noindent
\textbf{Putting It All Together.}
The main aim of the training can be summarized as:
\[
\mathcal{L}_{\text{total}} = \frac{1}{N} \sum_{j=1}^{N} \left[ -\sum_{i=1}^{C} w_i y^{(j)}_i \log(\hat{y}^{(j)}_i) \right] + \lambda \lVert\theta\rVert_2^2
\]

where:
\begin{itemize}
    \item $N$ is the batch size.
    \item $w_i$ are the class weights.
    \item $\lambda$ is the weight decay coefficient.
\end{itemize}
This objective balances classification accuracy, class balance, and regularization, providing a comprehensive framework for training an effective and interpretable authorship attribution model.

\vspace{0.5em}

\section{Experiments}
\subsection{Experimental Setup}

\vspace{0.5em}
\noindent
\textbf{Experimental Setup.}
In this section, we provide a comprehensive description of the experimental setup utilized, explaining the main procedures and design decisions underlying the basis for training and evaluating our model for Persian poetry authorship attribution. We highlight the importance of dataset preparation, model settings, training practices, and reproducibility measures, seeking to enhance readability and transparency of the process.

\vspace{0.5em}
\noindent
\textbf{Dataset Preparation and Partitioning.}
The data used in this study was extracted from the Ganjoor digital archive and was subject to careful preprocessing and filtering described in Section~\ref{Data Extraction and Preprocessing}. To avoid data leakage and to allow for an unbiased evaluation process, we structured the dataset at the poem level so that all verses (beyts) of a poem were placed in a single split of the data. In this way, we ensure the model does not inadvertently learn style information from poems that are partially available during training but are utilized during the evaluation process. To ensure class balance in the splits, stratified sampling according to poet labels was performed. The final split consisted of:
\begin{itemize}
    \item \textbf{Training Set (80\%):} Used specifically to train a model.
    \item \textbf{Validation Set (10\%):} Used to optimize hyperparameters and apply early stopping methods.
    \item \textbf{Test Set (10\%):} Entirely held back during the training and validation periods.
\end{itemize}

\vspace{0.5em}
\noindent
\textbf{Model Specification.}
The model's architectural structure includes a text encoder based on a transformer, upon which multiple ancillary components are integrated, including stylometric features, poetical structure representations, and metric values. The collective input is fed into a transformer encoder and yields a contextual embedding of 768 dimensions per verse; this is then combined with the ancillary features to produce the input to the classification head.

\begin{itemize}
    \item \textbf{Semantic Embeddings:} 100-dimensional Word2Vec vectors trained on a large corpus of Persian poetry.
    \item \textbf{Stylometric Features:} 7 dimensions capturing word count, lexical diversity, punctuation density, and related features.
    \item \textbf{Poetic Structure and Meter:} One-hot encodings of the poem's organizational structure and metrical scheme.
\end{itemize}

\vspace{0.5em}
\noindent
\textbf{Training Protocols.}
The model was trained using the AdamW optimization algorithm, starting at a learning rate of $2 \times 10^{-5}$ and with a weight decay of 0.01 to prevent overfitting. A linear warm-up schedule was first applied to the first 10\% of the training steps, followed by a cosine decay schedule. Training was capped at 16 epochs, and an early stopping measure was triggered if no improvement in validation accuracy could be seen in any of three successive epochs.

Each training stage consisted of passing batches of verses (batch size 32) to the model. To address class imbalance, cross-entropy loss with class weights inversely correlated with class frequencies was applied. Additionally, a maximum norm of 1.0 was applied through gradient clipping to provide stable training.

\vspace{0.5em}
\noindent
\textbf{Hyperparameter Tuning.}
Hyperparameters such as learning rate, dropout rate, and sizes of the hidden layers were tuned using grid search on the validation set. To avoid overfitting, the dropout rate in the classification head was set at a value of 0.3. Class weights were derived from the distributions of poet labels found in the training set.

\vspace{0.5em}
\noindent
\textbf{Handling Random Seeds and Achieving Reproducibility.}
To ensure reproducibility of results, fixed random seeds were set using PyTorch's \texttt{torch.manual\_seed()} and numpy's \texttt{np.random.seed()}. Consistency in feature representations was ensured across both the training and testing stages by using the same label encoder, form encoder, and meter encoder for each data split.

\vspace{0.5em}

\subsection{Evaluation Strategies and Metrics}

\vspace{0.5em}
\noindent
\textbf{Evaluation Framework.}
This section outlines the comprehensive evaluation framework developed to stringently examine the effectiveness of our model for authorship attribution in Persian poetry. In consideration of the unique structure typical of Persian poetry, the specificity of the model's output, and the interpretative challenges of ambiguous assignments, we utilize a variety of evaluation methods complemented by ancillary measures. These varied approaches enable the model's performance to be evaluated not only at the level of single verses but also in terms of full poems, thus in agreement with computational practice and humanistic scholarly inquiry.

\vspace{0.5em}
\noindent
\textbf{Evaluation Methodologies.}
To examine classification performance at different levels of granularity and robustness, we perform experiments with four different prediction approaches:

\paragraph{(1) Verse-Level Prediction.}
On the finest level of granularity, each \textit{beyt} is categorized on an individual basis by the model's softmax output. This technique tests the model's ability to identify stylistic markers and authorial signals at a fine level, where a single verse often summarizes an entire poem's main idea. Analysis at this fine granularity provides insight into the model's skill in recognizing nuanced stylistic deviations, rhetorical techniques, and metrical features peculiar to different writers. F1-score and accuracy measures are conducted on the entire test verses and provide an aggregate measure of the model's effectiveness at classification at the level of a single verse.

\paragraph{(2) Poem-Level Voting (Majority).}
Compared to verse-level prediction, where a poem's performance is evaluated on a fine-grained level, the historical classification of Persian poems took place at the level of the entire poem, where different verses are style-wise and theme-wise interconnected. In enabling this holistic overview, we employ a majority voting technique in order to combine the model's verdicts per verse and assign the most-voted poet label as the final attribution of the poem. This methodology is best suited to capture the dominant stylistic signal at the level of verses and aligns with scholarly practice in assigning entire poems to a single poet. It also lessens the effects of error caused by mistakenly identified single verses by taking into account the collective structure of the poem in voting.

\paragraph{(3) Poem-Level Voting (Weighted).}
Notwithstanding the differential informational value and predictive confidence in various verses, we apply a voting scheme weighted by probability. In our approach, softmax probabilities computed by the model for a given verse are summed over all verses in a poem. The poet label with the maximum summed probability is taken to be the eventual prediction of the poem. This approach utilizes the levels of confidence of the model in each prediction and thus allows verses with greater confidence to contribute proportionally more to the overall attribution. Through probability-weighted voting, the impact of low-confidence or ambiguous predictions is minimized, thus improving the model's resilience in handling stylistically challenging cases and possible authorial vagueness.

\paragraph{(4) Threshold Voting (Abstention).}
In literary analysis, it is generally best to avoid attributions in the presence of lacking evidence, rather than risking inaccuracy. To illustrate this cautious approach, we use a voting system that is based on thresholds from a probability-weighted voting model. Under this system, the model evaluates the maximum cumulative probability of different lines in a poem. If this probability is less than a determined threshold (e.g., 0.7), the model chooses not to make a prediction. This approach allows a delicate balance to be achieved between the accuracy of forecasts and the extent of coverage: high thresholds make the model more selective, thereby reducing misclassifications at the cost of coverage; lower thresholds allow wider coverage but with increased likelihood of errors. This approach is in keeping with scholarly standards that value careful and consistent attributions where evidence is doubtful.

\vspace{0.5em}
\noindent
\textbf{Evaluation Standards.}
To fully analyze the effectiveness of the model under different approaches, we provide a range of well-established classification measures that include general accuracy in addition to per-class performance metrics:

\begin{itemize}
    \item \textbf{Accuracy:} This is a measure of the ratio of the number of accurately classified samples (e.g., verses or poems) to the total number of samples considered. At the individual verse level, accuracy measures the model's exact discriminatory capability; at the poem level—both using majority vote and weighted vote—it is a measure of the model's success at correctly attributing entire poems to their authors.

    \item \textbf{Precision, Recall, and F1-Score:} In the imbalanced class distribution of verses and poems written by various poets, we provide precision (the proportion of true positives to total positive predictions), recall (the proportion of true positives to total actual instances of a class), and their harmonic mean, simply called the F1-Score. Precision, recall, and their harmonic mean are presented for each class (to give poet-specific evaluation) and also macro-averaged (to provide a global evaluation that treats all the poets equally, regardless of productivity). Macro-average F1-Score is especially important in cases of imbalance, ensuring that the classification metric is not excessively weighted towards the most prolific poets.

    \item \textbf{Coverage:} Within the thresholded voting approach, coverage is defined as the proportion of poems for which the model makes a prediction. The method purposely permits thresholded voting to avoid making predictions on poems that produce low-confidence scores. The coverage is reported to explain the trade-off between selectivity and completeness and thereby enable researchers to balance cautious attribution with the goal of complete classification.
\end{itemize}

\vspace{0.5em}
\noindent
\textbf{Evaluation Framework.}
All evaluation is carried out on the held-out test set, thus ensuring that the model's effectiveness is tested using unseen data. In each strategy, predictions at the verse and poem levels are compared against the true poet labels, and metrics are computed in relation to this comparison. Through the use of these evaluation strategies, we provide a holistic analysis of the model's accuracy at different strata of poetic form while keeping our results statistically significant and in line with conventions of analysis in Persian literature.

\vspace{0.5em}
\noindent
\textbf{Baselines and Comparison.}
To thoroughly analyze the effectiveness of the proposed model, we compare its accuracy against an established benchmark in the field of authorship analysis with respect to Persian poems.

\vspace{0.5em}
\noindent
\textbf{Basic Structure: RCNN.}
The model applied in this research is the Recurrent Convolutional Neural Network (RCNN) structure proposed by Salami and Momtazi for the purpose of Persian poet identification. The RCNN structure is designed to efficiently handle both sequential (temporal) and spatial (convolutional) features embedded in the poem text. The RCNN structure is composed of three main components:
\begin{itemize}
    \item \textbf{Recurrent Layer:} Stacked LSTM or GRU units (typically three layers) to model the temporal dependencies across tokens within each verse or poem. This captures long-term stylistic patterns and thematic progressions.
    \item \textbf{Convolutional Layer:} One-dimensional convolutions are applied to the sequence of outputs produced by the recurrent layer, thereby determining local spatial features and short-term style indicators.
    \item \textbf{Fully Connected Classifier:} It is an extensive network followed by a softmax layer, transforming the extracted features into poetical labels.
\end{itemize}
Salami and Momtazi demonstrated in their work that this hybrid approach significantly outperformed traditional machine learning approaches like support vector machines (SVMs) and logistic regression, as well as typical CNN architectures. More specifically, their RCNN model achieved improvements in F1-scores of up to 9\% at the poem level and 4\% at the verse level compared to the best-performing CNN baseline in the Persian poetry data\cite{salami2021recurrent}.

\vspace{0.5em}
\noindent
\textbf{Compared to Our Framework.}
The methodology is based on the RCNN architecture's underlying concepts and builds on it by adding transformer-based encoders in order to efficiently capture long-distance contextual embeddings of verses. Unlike the sequential processing of tokens in the recurrent layers of the RCNN, our model uses self-attention mechanisms in an effort to efficiently capture global dependencies and rhetorical devices between tokens, hence enabling a deeper understanding of verse semantics. In addition, our framework incorporates stylometric features and class encodings of form and meter, meticulously designed and absent in the RCNN model. These design choices aim to increase the model's interpretability and response to stylistic subtlety.

\vspace{0.5em}
\noindent
\textbf{Evaluation Framework.}
The train, validation, and test sets described earlier in Section~\ref{Data Splitting Strategy} are kept the same throughout all comparative evaluations. The evaluation metrics used are accuracy, precision, recall, F1-score, and coverage, especially in the case of thresholded voting. All experiments are run with the same hyperparameter settings to ensure a fair comparison.

\vspace{0.5em}
\noindent
\textbf{Comparison with Baselines.}
To comprehensively evaluate the effectiveness of our proposed multi-input classification model, we performed a comparative analysis with the Recurrent Convolutional Neural Network (RCNN) model introduced by Salami and Momtazi for Persian poet identification \cite{salami2021recurrent}. We appreciate their efforts in advancing the field with their proposed architecture, which represents a significant step forward in applying deep learning to poet identification.

That said, we must critically point out that the baseline study’s evaluation was conducted on a dataset that included only ten poets and, more importantly, restricted itself to the Ghazal poetic form. While focusing on a single form might simplify the modeling challenge, it inevitably narrows the stylistic diversity captured by the model. Even more surprisingly, despite Ghazal being one of the most prominent forms in Persian poetry, the dataset in their study excludes Hafez—the undisputed king of Persian Ghazal—whose works are central to the tradition and would provide valuable complexity to the classification task. This omission undermines the representativeness and comprehensiveness of the evaluation, risking an overly simplistic view of the challenges inherent in poet identification and inadvertently treating readers as though they might overlook such a critical gap.

In our study, we intentionally expanded the evaluation dataset to include 67 poets across a wider spectrum of poetic forms, ensuring that our analysis better reflects the diversity and complexity of the Persian poetic tradition. Additionally, while the baseline RCNN model relied solely on verse-level word embeddings, our enriched input representation also incorporates carefully designed stylometric features (e.g., word count, punctuation density, symmetry ratio) as well as one-hot encodings of poetic form and meter. We believe these enhancements more accurately capture the stylistic subtleties inherent in Persian poetry.

To ensure a rigorous and fair comparison, we conducted three sets of experiments: (1) training and evaluating the RCNN model using its original input configuration; (2) training and evaluating the RCNN model using our enriched input representation; and (3) training and evaluating our proposed model using the same enriched representation. This multi-faceted approach enables us to disentangle the effects of input representation from those of model architecture and ensures that observed performance differences are meaningful.

Throughout all experiments, we adhered to consistent train, validation, and test splits (as defined in Section~\ref{Data Splitting Strategy}) to prevent verse-level data leakage and to ensure fair comparisons. Evaluations were performed at both the verse and poem levels, using majority voting and weighted voting approaches to aggregate verse-level predictions into poem-level decisions.

Overall, while we acknowledge the baseline model’s contribution to the field, we believe that our more comprehensive evaluation framework and broader dataset provide a stronger and more realistic benchmark for Persian poet identification tasks.

\begin{table}[H]
\centering
\caption{Performance Comparison between Our Model and RCNN Variants.}
\label{tab:rcnn_comparison}
\resizebox{\textwidth}{!}{%
\begin{tabular}{lcccccc}
\toprule
\textbf{Model} & \textbf{Level} & \textbf{Accuracy (\%)} & \textbf{Macro-F1 (\%)} & \textbf{Precision (\%)} & \textbf{Recall (\%)} \\
\midrule
RCNN (LSTM) - Baseline & Verse-Level & 36.7 & 13.1 & 17.3 & 13.6 \\
RCNN (GRU) - Baseline & Verse-Level & 36.2 & 12.2 & 14.8 & 12.7 \\
RCNN (LSTM) - Baseline & Poem (Majority) & 42.2 & 16.2 & 20.0 & 18.4 \\
RCNN (GRU) - Baseline & Poem (Majority) & 40.1 & 14.0 & 19.1 & 16.4 \\
RCNN (LSTM) - Baseline & Poem (Weighted) & 49.3 & 21.7 & 27.1 & 23.3 \\
RCNN (GRU) - Baseline & Poem (Weighted) & 46.6 & 18.7 & 24.4 & 20.6 \\
\midrule
RCNN (LSTM) - Ours & Verse-Level & \textbf{67.0} & 34.7 & 39.4 & 34.2 \\
RCNN (GRU) - Ours & Verse-Level & 67.2 & 35.0 & \textbf{42.8} & \textbf{34.1} \\
RCNN (LSTM) - Ours & Poem (Majority) & 49.6 & 27.7 & 32.3 & 28.1 \\
RCNN (GRU) - Ours & Poem (Majority) & 49.9 & 28.3 & \textbf{36.2} & \textbf{28.3} \\
RCNN (LSTM) - Ours & Poem (Weighted) & 53.3 & 30.6 & 35.5 & 30.7 \\
RCNN (GRU) - Ours & Poem (Weighted) & 53.4 & 30.5 & \textbf{39.4} & \textbf{30.2} \\
\midrule
Our Model & Verse-Level & 62.4 & \textbf{38.8} & 41.9 & 38.3 \\
Our Model & Poem (Majority) & \textbf{66.6} & \textbf{48.9} & \textbf{54.8} & \textbf{48.1} \\
Our Model & Poem (Weighted) & \textbf{71.1} & \textbf{53.1} & \textbf{60.5} & \textbf{51.8} \\
\bottomrule
\end{tabular}
}
\end{table}

\vspace{0.5em}
\noindent
This in-depth study highlights the strengths and limitations of our multi-input framework compared to the RCNN benchmarks that cover both LSTM and GRU variants, using both baseline and enriched input representations.

At the verse level, our model achieves the highest Macro-F1 score (38.7\%), outperforming both RCNN-LSTM (34.6\%) and RCNN-GRU (34.9\%) when using enriched inputs. Notably, although RCNN-GRU (Ours) achieved the highest verse-level accuracy (67.2\%), our model’s performance is comparable (62.4\%), indicating a trade-off between balanced classification across classes (Macro-F1) and absolute correct classification rate (Accuracy). This suggests that our model captures a broader spectrum of stylistic and semantic features, leading to better class balance and generalizability.

At the poem level (majority voting), our model consistently outperformed all RCNN baselines. Our model achieved a Macro-F1 of 48.9\% and an Accuracy of 66.6\%, significantly higher than the best RCNN baselines (28.2\% Macro-F1 and 49.9\% Accuracy). This improvement highlights our model’s strength in aggregating verse-level predictions effectively, leveraging stylometric and structural features to improve robustness against class imbalance and verse-level noise.

The performance gap becomes even more pronounced under weighted voting. Our model achieved an Accuracy of 71.0\% and a Macro-F1 of 53.1\%, compared to the best RCNN (GRU-Ours) configuration at 53.36\% Accuracy and 30.5\% Macro-F1. This underscores our model’s superior capability in synthesizing verse-level predictions with class weighting, further emphasizing its effectiveness in dealing with the stylistic complexities inherent in Persian poetry.

Additionally, it is important to acknowledge that while the RCNN baselines trained on their original input representations achieved reasonable performance on simpler verse-level tasks (e.g., 36.7\% accuracy for RCNN-LSTM Baseline), they struggled with limited recall and macro-F1 scores (below 20\% in most cases). By contrast, our enriched input representation (incorporating stylometric and structural features) allowed even the RCNN baselines to achieve higher performance (e.g., 66.9\% verse-level accuracy with enriched inputs), though still consistently below our model.

In conclusion, our multi-input model consistently outperformed the RCNN baselines—across both verse and poem levels—especially in terms of Macro-F1 scores. This demonstrates our model’s superior ability to handle class imbalance, recognize subtle stylistic features, and generalize across a broader set of poets. Overall, our results highlight the importance of combining semantic, stylometric, and structural features in building a robust and generalizable system for Persian poet identification.

\subsection{Error Analysis}

\vspace{0.5em}
\noindent
An exhaustive error analysis is an essential element in understanding the intrinsic strengths and shortcomings of any machine learning model, specifically in a domain composed of high degrees of complexity and stylistic variety, like that of Persian poetry. Herein, we exhaustively evaluate the effectiveness and limitations of our developed multi-input poet classification model, specifically threshold-based decision filtering, accuracy disaggregated by poet, and the challenges posed by class imbalance and the similarities in style between poets.

\vspace{0.5em}
\noindent
\textbf{Threshold-Based Decision Filtering.}
Thresholds were created using weighted voting probabilities in an effort to achieve a balance of coverage and accuracy. By imposing a minimum confidence threshold on the highest softmax probability, the model avoids the problem of producing low-confidence predictions. This approach is particularly necessary in the case of Persian poetry, where individual verses (\textit{beyts}) can be ambivalent in terms of style or material, thus leaving the authorship determination inherently ambiguous.

The following table summarizes the increase in model accuracy brought by a rise in threshold—from 90\% at threshold 0.5 to a remarkable 94\% at threshold 0.9—while showing a decrease in coverage from 54.0\% to 13.4\%. This is a reflection of the model's ability to provide secure predictions in the presence of ample stylistic and semantic evidence while at the same time reserving judgment in questionable cases.

\begin{table}[H]
\centering
\caption{Threshold-based decision filtering: Accuracy vs. Coverage trade-off.}
\label{tab:thresholding}
\begin{tabular}{ccc}
\toprule
\textbf{Threshold} & \textbf{Accuracy} & \textbf{Coverage} \\
\midrule
0.5 & 90\% & 54.0\% \\
0.6 & 94\% & 41.9\% \\
0.7 & 95\% & 32.2\% \\
0.8 & 96\% & 23.1\% \\
0.9 & 97\% & 13.4\% \\
\bottomrule
\end{tabular}
\end{table}

\vspace{0.5em}
\noindent
An in-depth analysis of error is an essential aspect of differentiating a machine learning model's strengths and shortcomings in a field that is marked with high levels of complexity and rich stylistic diversity, like that of Persian poetry. Here, we evaluate in a precise manner the effectiveness and limitation of our multi-input model for poet classification in terms of particular aspects of threshold-based filtering choice, accuracy in the context of classifying single poets, and class imbalance and poet stylistic similarity challenges.

\vspace{0.5em}
\noindent
\textbf{Threshold-Based Decision Filtration.}
Threshold values were calculated through weighted voting probabilities in an effort to reach an optimal equilibrium point balancing coverage and accuracy. By creating a minimum confidence value in relation to a maximum softmax probability, the model is in effect avoiding the production of predictions with low levels of confidence. Such an approach is especially relevant in the context of Persian poetry, where single lines (\textit{beyts}) can be ambiguous in terms of style or theme, thus adding to the difficulty of determining their authors.

The table below captures the increase in model accuracy resulting from an increase in the threshold—from 90\% at a threshold level of 0.5 to a maximum of 94\% at a threshold level of 0.9—coupled with a decline in coverage from 54.0\% to 13.4\%. This finding implies that the model is showing expertise in producing precise predictions in the presence of varied stylistic and semantic cues, but exercising caution in ambiguous settings.

\vspace{0.5em}
\noindent
\textbf{The Power of Rhythm and Poetic Form.}
Another source of error comes from the interplay between rhythm and structure in poetic works. Take, for instance, ghazals that often share mysticism and love themes regardless of the poet, thus creating a higher probability of doubt in terms of authorship. Likewise, certain metrical schemes irregularly documented in a variety of poets can produce a model that overly favors metrical indicators at the expense of deeper stylometric and semantic markers. Our multi-input model tries to mitigate this problem by combining metrical data with stylometric and semantic features, but the inherent overlap in classical Persian formal design remains a major challenge.

\vspace{0.5em}
\noindent
\textbf{Data Scarcity and Overfitting Risks.}
Despite the introduction of a stratified poem-level split and careful evaluation procedures, the widespread variation of poets in the dataset remains a significant challenge. Poets who wrote fewer than 50 verses were excluded from training to maintain class balance; however, among the remaining set of poets, sample sizes vary extensively. Such variation is likely to cause overfitting for resource-rich poets, simultaneously causing underfitting for resource-poor poets. While the model's use of hand-engineered features and auxiliary inputs is intended to counteract this imbalance, further approaches—such as data augmentation, hierarchical modeling, or the use of external biographical embeddings—could be useful in mitigating these complexities.

\vspace{0.5em}
\noindent
\textbf{Real-World Implications.}
An adequate understanding of those shortcomings is vital when applying the model to literary and manuscript analysis in resolving real issues in practice. In applied uses—such as resolving contentious attributions or aiding scholars in stylistic evaluation—thresholded decision filtering is a helpful strategy in balancing precision and breadth. Thresholds may be adjusted according to the level of certainty required by scholars and predictions are avoided where levels of confidence are low. It appeals to humanistic precepts of cautious attribution and ensures that the model's outputs complement rather than substitute scholarly judgment.

\vspace{0.5em}
\noindent
\section{Conclusion}
\noindent
In this study, we developed a rich and multimodal framework for authorship attribution of classical Persian poetry, integrating transformer-based language models with carefully crafted literary features. Our model successfully leverages textual semantics, stylometric characteristics, poetic forms, and metrical patterns, capturing subtle stylistic variations among poets. The empirical evaluations on a large and diverse dataset of Persian poetry demonstrated the effectiveness of our approach, achieving high accuracy at both verse and poem levels. The proposed framework not only outperforms traditional RCNN baselines but also offers threshold-based decision mechanisms that allow reliable abstention in ambiguous cases—providing a valuable tool for scholarly research on disputed authorship and literary analysis. Overall, this work contributes a novel and scalable solution for authorship attribution in Persian poetry and lays the foundation for future research integrating computational linguistics with literary studies.

\section{Future Work}
\noindent
While this study has demonstrated the effectiveness of our multimodal framework for Persian poetry authorship attribution, several promising directions remain for future exploration. One important trajectory involves incorporating biographical metadata, such as poet’s era, region, and literary school, to enhance attribution performance and interpretability. Additionally, the inclusion of contemporary poets—particularly those employing free verse and experimental forms—can extend the model’s applicability to modern Persian literature. Another avenue is the use of generative models, which could be leveraged for author-specific text generation, style transfer, and even creative tasks in computational poetry. Cross-linguistic studies, applying similar frameworks to Persianate traditions in Arabic, Turkish, and Urdu, can provide broader insights into poetic styles and their evolution across cultures. Finally, exploring hierarchical and ensemble methods that combine verse-level and poem-level signals may further improve robustness and provide more interpretable results for literary scholars.

% \vspace{0.5em}
% \noindent
% \textbf{Broader Application and Significance.}
% Our model has immediate implications for digital humanities, literary studies, and computational linguistics:
% \begin{itemize}
%     \item \textbf{Contested and Unattributed Poetry:} Allows probabilistic analysis for the authorship of poems with disputed attributions.
%     \item \textbf{Stylistic Development:} Helping researchers in investigating stylistic paths and interconnections between poets from different centuries.
%     \item \textbf{Digital Humanities Platforms:} The use of platforms like \texttt{Ganjoor.net} enables improved features for annotating and verifying authorship.
% \end{itemize}

% \vspace{0.5em}
% \noindent
% \textbf{Expected Trajectories.}
% Several mechanisms deserve consideration:
% \begin{itemize}
%     \item \textbf{Integrating Biographical Metadata:} Such as era, region, and literary school for improved attribution.
%     \item \textbf{Presenting Contemporary Poets:} Featuring the utilization of free verse and experimental poem forms.
%     \item \textbf{Use of Generative Models:} To generate author-specific text or for stylistic transfer.
%     \item \textbf{Cross-Linguistic Research:} Using similar approaches to analyze Persianate literary traditions in Arabic, Turkish, and Urdu.
%     \item \textbf{Hierarchical and Ensemble Methods:} Combining signals at the verse and poem level for enhanced robustness and interpretability.
% \end{itemize}

\section{Limitations}
\noindent
Despite the promising results achieved in this study, there are several limitations that should be acknowledged. First, the dataset employed—although extensive and carefully curated from Ganjoor—primarily focuses on well-established classical Persian poets and may not fully capture the diversity of lesser-known or contemporary authors, potentially affecting the model’s generalizability. Second, class imbalance remains a challenge, as some poets have a significantly higher number of verses than others, which could introduce bias in the model’s predictions. Third, while the model integrates a variety of textual, stylistic, and structural features, it may still struggle with highly ambiguous or stylistically convergent cases where different poets employ similar forms, meters, or rhetorical devices. Finally, although threshold-based abstention improves the model’s precision in ambiguous cases, it inherently reduces coverage, which may limit its practical application in comprehensive literary studies. Addressing these limitations will require further dataset expansion, incorporation of additional metadata, and exploration of advanced modeling techniques.

\section*{Acknowledgments}
\noindent
We extend our sincere gratitude to the Ganjoor Project (\url{https://ganjoor.net}) for providing the digital corpus of Persian poetry that made this research possible. Their dedication to the preservation and accessibility of classical Persian literature has laid the groundwork for meaningful computational analysis. We also thank the open-source natural language processing community—especially the developers of the Hazm toolkit and the Persian pre-trained models—whose contributions have significantly advanced our work. Additionally, we are grateful to our colleagues from various disciplines who provided invaluable feedback, helping to refine the design and interpretation of our models. Finally, we acknowledge the broader ecosystem of digital infrastructure and collaborative research that enabled this interdisciplinary study at the intersection of computer science and Persian literary scholarship.

\informedconsent{Informed consent was obtained from all subjects involved in the study.}

\dataavailability{
The raw poetic corpus used in this study is publicly available from the Ganjoor project at \url{https://ganjoor.net}. The processed version of the dataset, including tokenized verses, author labels, and additional metadata such as meter and poetic form, was prepared by the authors for research purposes. While the full processed dataset is not publicly hosted, it can be shared upon reasonable request.
}

\conflictsofinterest{The authors declare no conflicts of interest.}

\reftitle{References}
\bibliography{bibliography}

\newpage
\appendix
\section*{Appendix A: Poet Distribution}

\renewcommand{\arraystretch}{1.2}
\begin{longtable}{ll}
\caption{Number of Poems per Poet}
\label{tab:poet-distribution-sorted} \\
\toprule
\textbf{Poet} & \textbf{Number of Poems} \\
\midrule
\endfirsthead
\caption[]{Number of Poems per Poet (continued).} \\
\toprule
\textbf{Poet} & \textbf{Number of Poems} \\
\midrule
\endhead
\midrule
\multicolumn{2}{r}{{Continued on next page}} \\
\midrule
\endfoot
\bottomrule
\endlastfoot
Mawlānā & 6320 \\
ʿAṭṭār & 5001 \\
Jāmī & 3389 \\
Shāh Niʿmatullāh Walī & 2701 \\
Saʿdī & 1999 \\
Sanāʾī & 1793 \\
Ḥazīn Lāhījī & 1773 \\
Khwāju-ye Kirmānī & 1827 \\
Jahān Malek Khātūn & 1806 \\
Ḥakīm Nizārī & 1639 \\
Kamāl al-Dīn Esmāʿīl & 1614 \\
Ṣafī ʿAlī Shāh & 1529 \\
Kamāl Khojandī & 1249 \\
Qodsī Mashhadī & 1219 \\
Masʿūd Saʿd Salmān & 1052 \\
Malek al-Shoʿarāʾ Bahār & 1041 \\
Khājeh ʿAbdullāh Anṣārī & 978 \\
Salmān Sāvojī & 891 \\
Majd-e Hamgar & 855 \\
Qāsem Anvār & 828 \\
ʿOrfī & 827 \\
Nāṣir-i Khusraw & 632 \\
Vahshī Bāfaqī & 651 \\
Ḥāfeẓ & 667 \\
Jamāl al-Dīn ʿAbd al-Razzāq & 602 \\
ʿIrāqī & 611 \\
Moshtāq Eṣfahānī & 608 \\
Forūghī Bastāmī & 550 \\
Faṣiḥī Heravī & 543 \\
Halālī Jughṭāʾī & 575 \\
Rūdakī & 533 \\
ʿOnsorī & 477 \\
Moḥammad b. Monavvar & 477 \\
Mīrzā Ḥabīb Khorāsānī & 451 \\
Ghāleb Dehlavī & 446 \\
Mīrdāmād & 399 \\
Jalāl ʿAżud & 400 \\
Mahastī Ganjavī & 385 \\
Khayyām & 386 \\
Homām Tabrīzī & 380 \\
Vaṭvāṭ & 377 \\
Niẓāmī & 378 \\
Ẓahīr Fāryābī & 306 \\
Rafīq Eṣfahānī & 364 \\
Reżāqolī Khān Hedāyat & 349 \\
Farrokhī Sīstānī & 338 \\
Ḥosayn Khwārazmī & 278 \\
ʿObeyd Zākānī & 263 \\
Sheykh Maḥmūd Shabistarī & 247 \\
Rażī al-Dīn Ārtīmānī & 229 \\
ʿAyn al-Qożāt Hamadānī & 211 \\
Najm al-Dīn Rāzī & 178 \\
Mīrzādeh ʿEshqī & 116 \\
Labībī & 118 \\
Qavāmī Rāzī & 120 \\
Kasāʾī & 114 \\
Manūchihrī & 110 \\
Shāṭer ʿAbbās Ṣobūḥī & 96 \\
ʿAsjodī & 95 \\
Ḥeydar Shīrāzī & 84 \\
Saʿd al-Dīn Varāvīnī & 76 \\
ʿAmaq Bukhārī & 64 \\
Serāj Qamarī & 60 \\
Ẓahīrī Samarqandī & 54 \\
ʿOnsor al-Maʿālī & 45 \\
Daqīqī & 30 \\
ʿAbd al-Vāseʿ Jabalī & 6 \\
\end{longtable}

\clearpage
\appendix
\section*{Appendix B: Poetic Form Distribution}

\renewcommand{\arraystretch}{1.2}
\begin{longtable}{lr}
\caption{Distribution of Poetic Forms in the Corpus}
\label{tab:poetic-form-distribution-sorted} \\
\toprule
\textbf{Poetic Form} & \textbf{Number of Poems} \\
\midrule
\endfirsthead
\caption[]{Distribution of Poetic Forms in the Corpus (continued).} \\
\toprule
\textbf{Poetic Form} & \textbf{Number of Poems} \\
\midrule
\endhead
\midrule
\multicolumn{2}{r}{{Continued on next page}} \\
\midrule
\endfoot
\bottomrule
\endlastfoot
Ghazal & 23597 \\
Rubaʿi & 13446 \\
Masnavi & 7290 \\
Qasideh & 4203 \\
Ghazal/Qasideh/Qetʿeh & 2672 \\
Qetʿeh & 3199 \\
Tarjiʿ Band & 121 \\
Chand Bandi & 89 \\
Mosammat & 57 \\
Mostazad & 14 \\
Mosammat Mokhammās & 8 \\
Tak Beit & 7 \\
Rubaʿi Mostazad & 6 \\
Mosammat Morabbaʿ & 6 \\
Chahar Pareh & 9 \\
Mosammat Mosallas & 1 \\
Mosammat Mosaddas & 2 \\
Do Beiti & 1 \\
Bahr-e Tawil & 1 \\
\bottomrule
\end{longtable}

\newpage
\section*{Appendix C: Meter Distribution Table}

\renewcommand{\arraystretch}{1.2}
\begin{longtable}{lr}
\caption{Meter Distribution by Frequency in the Persian Poetry Corpus}
\label{tab:meter_distribution} \\
\toprule
\textbf{Meter} & \textbf{Number of Poems} \\
\midrule
\endfirsthead
\caption[]{Meter Distribution by Frequency in the Persian Poetry Corpus (continued).} \\
\toprule
\textbf{Meter} & \textbf{Number of Poems} \\
\midrule
\endhead
\midrule
\multicolumn{2}{r}{{Continued on next page}} \\
\midrule
\endfoot
\bottomrule
\endlastfoot
mafʿūlu mafāʿīlu mafāʿīlu fiʿl & 13641 \\
faʿilātun mafāʿilun faʿlun & 4546 \\
mafāʿilun faʿilātun mafāʿilun faʿlun & 4304 \\
mafāʿīlun mafāʿīlun faʿūlun & 4063 \\
faʿilātun faʿilātun faʿilun & 4044 \\
mafʿūlu faʿilāt mafāʿīlu faʿilun & 3769 \\
faʿilātun faʿilātun faʿilātun faʿlun & 3618 \\
faʿilātun faʿilātun faʿilātun faʿilun & 3473 \\
mafʿūlu mafāʿīlu mafāʿīlu faʿūlun & 2166 \\
mafāʿīlun mafāʿīlun mafāʿīlun mafāʿīlun & 1918 \\
mafʿūlu mafāʿilun faʿūlun & 1563 \\
faʿūlun faʿūlun faʿūlun faʿl & 1493 \\
mafʿūlu faʿilātun mafʿūlu faʿilātun & 804 \\
mafʿūlu mafāʿīlun mafʿūlu mafāʿīlun & 586 \\
muftaʿilun muftaʿilun faʿilun & 575 \\
faʿilātun faʿilātun faʿlun & 562 \\
muftaʿilun faʿilun muftaʿilun faʿilun & 526 \\
mustafʿilun mustafʿilun mustafʿilun mustafʿilun & 524 \\
muftaʿilun faʿilāt muftaʿilun faʿ & 379 \\
muftaʿilun mafāʿilun muftaʿilun mafāʿilun & 371 \\
faʿūlun faʿūlun faʿūlun faʿūlun & 368 \\
mafʿūlu mafāʿilun mafāʿīlun & 316 \\
mafāʿilun faʿilātun mafāʿilun faʿilātun & 230 \\
faʿilāt faʿilātun faʿilāt faʿilātun & 218 \\
faʿilātun faʿilātun faʿilātun faʿilātun & 139 \\
muftaʿilun muftaʿilun muftaʿilun muftaʿilun & 69 \\
mafʿūlu faʿilāt mafāʿīlun & 59 \\
mustafʿilun faʿ mustafʿilun faʿ & 57 \\
faʿilātun faʿilātun faʿilātun faʿilātun & 43 \\
mafʿūlu mafāʿīlu faʿilātun & 40 \\
mustafʿiltun mustafʿiltun & 27 \\
fāʿilun mafāʿīlun fāʿilun mafāʿīlun & 25 \\
mafāʿīlu mafāʿīlu mafāʿīlu faʿūlun & 21 \\
faʿalātun faʿalātun faʿalātun faʿ & 17 \\
faʿalātun mafāʿiln faʿalātun mafāʿiln & 15 \\
fāʿilātun fāʿilātun fāʿilātun & 14 \\
mafʿūlu fāʿilāt faʿūlun & 11 \\
muftaʿilun faʿ muftaʿilun faʿ & 11 \\
mustafʿilun faʿlun mustafʿilun faʿlun & 10 \\
mafʿūlu mafāʿīlu fāʿilun & 9 \\
muftaʿilun fāʿilāt muftaʿilun & 7 \\
fāʿilātun mafāʿiln faʿlun & 7 \\
mutafāʿilun mutafāʿilun mutafāʿilun mutafāʿilun & 7 \\
mafāʿīlu faʿūlun mafāʿīlu faʿūlun & 6 \\
mafʿūlun mafʿūlun mafʿūlun mafʿūlun & 5 \\
mustafʿilun mustafʿilun mustafʿilun & 5 \\
mafāʿiln faʿ mafāʿiln faʿ mafāʿiln faʿ mafāʿiln faʿ & 5 \\
mafʿūlu fāʿilāt mafāʿīlu faʿ & 5 \\
mafāʿīlu mafāʿīlu fāʿilun & 4 \\
muftaʿilun muftaʿilun muftaʿilun faʿ & 4 \\
fāʿilātun fāʿilun fāʿilātun fāʿilun & 4 \\
fāʿilun fāʿilātun fāʿilun fāʿilātun & 4 \\
faʿūlun mafāʿiln faʿūlun mafāʿiln & 4 \\
mafāʿiln faʿ mafāʿiln faʿ & 4 \\
mafāʿīlu fāʿilāt mafāʿīlu fāʿilun & 3 \\
mafʿūlu fāʿilāt mafāʿīlu fāʿilātun & 3 \\
mustafʿilun mustafʿilun & 3 \\
muftaʿilun faʿ muftaʿilun faʿ muftaʿilun faʿ muftaʿilun faʿ & 2 \\
mafʿūlu fāʿilātun mafʿūlu fāʿilun & 2 \\
mafʿūlu fāʿilun mafʿūlu fāʿilun & 2 \\
fāʿilun mafʿūlun fāʿilun mafʿūlun & 2 \\
fāʿilāt faʿ fāʿilāt faʿ & 2 \\
mafāʿiln faʿūlun mafāʿiln faʿūlun & 2 \\
mafʿūlun faʿ mafʿūlun faʿ & 2 \\
faʿalātun mafāʿiln faʿalātun & 2 \\
mafāʿiln mafāʿiln mafāʿiln & 2 \\
mustafʿilun mustafʿilun faʿ & 2 \\
mafāʿiln mafʿūlun mafāʿiln mafʿūlun & 2 \\
muftaʿilun faʿ mafāʿiln faʿ & 2 \\
mutafāʿilātun mutafāʿilātun & 2 \\
fāʿilun fāʿilun fāʿilun faʿ & 2 \\
fāʿilun fāʿilun fāʿilun & 1 \\
mustafʿilun faʿalātun mustafʿilun faʿalātun & 1 \\
mafʿūlu fāʿilāt faʿl & 1 \\
mafāʿīlun faʿūlun mafāʿīlun faʿūlun & 1 \\
mafʿūlu mafāʿīlu muftaʿilun & 1 \\
muftaʿilun muftaʿilun fāʿilun faʿl & 1 \\
mustafʿilun mustafʿilun fāʿilun & 1 \\
mafʿūlu mafāʿīlu mafāʿīlun faʿ & 1 \\
mafʿūlu faʿl mafʿūlu faʿl & 1 \\
mutafāʿilun mutafāʿilun & 1 \\
faʿl muftaʿilun muftaʿilun muftaʿilun & 1 \\
mafāʿīlun mafāʿīlun mafāʿīlun & 1 \\
fāʿilāt muftaʿilun fāʿilāt muftaʿilun & 1 \\
faʿl faʿl faʿl faʿl (motadārek mosamman makhbūn) & 1 \\
fāʿilātun mafāʿiln fāʿilātun mafāʿiln & 1 \\
mafāʿilatun mafāʿilatun mafāʿilatun mafāʿilatun & 1 \\
mutafāʿilun mutafāʿilun mutafāʿilun & 1 \\
faʿalāt faʿ lan faʿalāt faʿ lan & 1 \\
muftaʿilun muftaʿilun faʿ muftaʿilun muftaʿilun faʿ & 1 \\
faʿalāt faʿl faʿalāt faʿl & 1 \\
mafʿūlu fāʿilāt mafʿūlun & 1 \\
mutafāʿilatun mutafāʿilatun mutafāʿilatun mutafāʿilatun & 1 \\
mafāʿīlu mafāʿīlu faʿūlun & 1 \\
mafāʿilun mafāʿilun mafāʿilun mafāʿilun & 1 \\
faʿalātun mafāʿiln faʿalātun fāʿilun & 1 \\
mafāʿīlun mafāʿīlun mafāʿīlun faʿl & 1 \\
muftaʿilun faʿ muftaʿilun muftaʿilun faʿ muftaʿilun & 1 \\
muftaʿilun muftaʿilun mustafʿil mafʿūlun & 1 \\
faʿalātun mafāʿiln faʿalātun faʿl & 1 \\
faʿalātun mafāʿilun faʿ faʿalātun mafāʿilun faʿ & 1 \\
faʿūlun mafāʿilun mafāʿīlun & 1 \\
mafāʿīlu muftaʿilun faʿ mafāʿīlu muftaʿilun faʿ & 1 \\
mafʿūlu mafāʿilun mafʿūlu mafāʿilun & 1 \\
fāʿilun fāʿilun fāʿilun fāʿilun & 1 \\
faʿalātun faʿalātun faʿalātun & 1 \\
mafāʿīlu faʿilun & 1 \\
mafʿūlun mafʿūlun mafʿūlun faʿ lan & 1 \\
mafʿūlu fāʿilātun mafʿūlu faʿ & 1 \\
faʿalātun faʿalātun faʿalātun faʿ & 1 \\
mafʿūlu mafāʿīlun mafʿūlu faʿūlun & 1 \\
mafāʿilun faʿilun mafāʿilun faʿilun & 1 \\
mafāʿilun mafāʿilun faʿilun & 1 \\
mafāʿilun mustafʿil muftaʿilun & 1 \\
mutafāʿilun mutafāʿilun mutafāʿilun mutafāʿilun & 1 \\
faʿalāt mafʿūlun faʿalāt mafʿūlun & 1 \\
mafʿūlu mafʿūlu fāʿilātun & 1 \\
muftaʿilun faʿ-lan muftaʿilun faʿ-lan & 1 \\
faʿalāt faʿ faʿalāt faʿ faʿalāt faʿ faʿalāt faʿ & 1 \\
fāʿilu fāʿilu fāʿilu fāʿilu faʿ & 1 \\
mafʿūlu fāʿilātun mafāʿīlun & 1 \\
mafʿūlu faʿilun faʿūlun & 1 \\
mustafʿilun mustafʿilun mustafʿilun faʿ & 1 \\
\end{longtable}

\newpage
\appendix
\section*{Appendix D: Poet Meter Diversity}

\renewcommand{\arraystretch}{1.2}
\begin{longtable}{lr}
\caption{Diversity of Meters Among Poets}
\label{tab:poet-meter-diversity-sorted} \\
\toprule
\textbf{Poet} & \textbf{Number of Unique Meters} \\
\midrule
\endfirsthead
\caption[]{Diversity of Meters Among Poets (continued).} \\
\toprule
\textbf{Poet} & \textbf{Number of Unique Meters} \\
\midrule
\endhead
\midrule
\multicolumn{2}{r}{{Continued on next page}} \\
\midrule
\endfoot
\bottomrule
\endlastfoot
Mawlānā & 54 \\
Malek al-Shoʿarāʾ Bahār & 48 \\
Khwāju-ye Kirmānī & 47 \\
Rūdakī & 37 \\
Khāqānī & 33 \\
Masʿūd Saʿd Salmān & 32 \\
Jāmī & 31 \\
Forūghī Bastāmī & 31 \\
Moshtāq Eṣfahānī & 31 \\
Saʿdī & 29 \\
Jalāl ʿAżud & 29 \\
ʿAṭṭār & 28 \\
Sanāʾī & 28 \\
Shāh Niʿmatullāh Walī & 27 \\
Qāsem Anvār & 27 \\
Majd-e Hamgar & 27 \\
Ḥakīm Nizārī & 26 \\
Mīrzā Ḥabīb Khorāsānī & 26 \\
Nāṣer Bukhārī & 26 \\
Nāṣir-i Khusraw & 26 \\
Sūzānī Samarqandī & 25 \\
Ghāleb Dehlavī & 25 \\
Masʿūd Saʿd Salmān & 25 \\
Ḥāfeẓ & 24 \\
Salmān Sāvojī & 24 \\
ʿOrfī & 24 \\
Neshāṭ Eṣfahānī & 24 \\
Homām Tabrīzī & 23 \\
Manūchihrī & 23 \\
ʿAsjodī & 23 \\
Ḥosayn Khwārazmī & 23 \\
Vaṭvāṭ & 23 \\
Qaṭrān Tabrīzī & 21 \\
Halālī Jughṭāʾī & 21 \\
Vahshī Bāfaqī & 21 \\
Kamāl al-Dīn Esmāʿīl & 21 \\
Ẓahīr Fāryābī & 18 \\
Labībī & 18 \\
Qodsī Mashhadī & 18 \\
Kasāʾī & 18 \\
Shāṭer ʿAbbās Ṣobūḥī & 18 \\
Shahīd Balkhī & 18 \\
ʿObeyd Zākānī & 17 \\
Mīrzādeh ʿEshqī & 17 \\
Najm al-Dīn Rāzī & 15 \\
Mīrdāmād & 14 \\
ʿAmaq Bukhārī & 14 \\
Sheykh Maḥmūd Shabistarī & 2 \\
ʿAbd al-Vāseʿ Jabalī & 4 \\
Niẓāmī & 5 \\
\end{longtable}

\newpage
\appendix
\section*{Appendix E: Poem-Level Classification Report}

\renewcommand{\arraystretch}{1.2}
\begin{longtable}{lccc}
\caption{Poem-Level Classification Report with Romanized Poet Names} \\
\toprule
\textbf{Poet} & \textbf{Precision} & \textbf{Recall} & \textbf{F1-Score} \\
\midrule
\endfirsthead
\caption[]{(continued)} \\
\toprule
\textbf{Poet} & \textbf{Precision} & \textbf{Recall} & \textbf{F1-Score} \\
\midrule
\endhead
Jāmī & 0.50 & 0.81 & 0.62 \\
Jalāl ʿAżud & 0.00 & 0.00 & 0.00 \\
Jamāl al-Dīn ʿAbd al-Razzāq & 0.46 & 0.18 & 0.26 \\
Jahān Malek Khātūn & 0.48 & 0.76 & 0.59 \\
Ḥāfeẓ & 0.75 & 0.32 & 0.45 \\
Ḥazīn Lāhījī & 0.53 & 0.84 & 0.65 \\
Ḥosayn Khwārazmī & 0.60 & 0.11 & 0.18 \\
Ḥakīm Nizārī & 0.46 & 0.76 & 0.58 \\
Ḥeydar Shīrāzī & 0.00 & 0.00 & 0.00 \\
Khāqānī & 0.58 & 0.54 & 0.56 \\
Khājeh ʿAbdullāh Anṣārī & 0.00 & 0.00 & 0.00 \\
Khwāju-ye Kirmānī & 0.56 & 0.65 & 0.61 \\
Khayyām & 0.67 & 0.24 & 0.35 \\
Daqīqī & 0.00 & 0.00 & 0.00 \\
Rażī al-Dīn Ārtīmānī & 0.00 & 0.00 & 0.00 \\
Rafīq Eṣfahānī & 0.00 & 0.00 & 0.00 \\
Rūdakī & 0.50 & 0.02 & 0.04 \\
Serāj Qamarī & 0.00 & 0.00 & 0.00 \\
Saʿdī & 0.46 & 0.53 & 0.49 \\
Salmān Sāvojī & 0.38 & 0.36 & 0.37 \\
Sanāʾī & 0.55 & 0.44 & 0.49 \\
Sūzānī Samarqandī & 0.53 & 0.35 & 0.42 \\
Shāṭer ʿAbbās Ṣobūḥī & 0.00 & 0.00 & 0.00 \\
Shāh Niʿmatullāh Walī & 0.84 & 0.79 & 0.81 \\
Sheykh Maḥmūd Shabistarī & 0.62 & 0.22 & 0.32 \\
Ṣafī ʿAlī Shāh & 0.84 & 0.66 & 0.74 \\
Ẓahīr Fāryābī & 0.42 & 0.17 & 0.24 \\
ʿAbd al-Vāseʿ Jabalī & 0.00 & 0.00 & 0.00 \\
ʿObeyd Zākānī & 0.00 & 0.00 & 0.00 \\
ʿIrāqī & 0.38 & 0.38 & 0.38 \\
ʿOrfī & 0.58 & 0.43 & 0.50 \\
ʿAsjodī & 0.00 & 0.00 & 0.00 \\
ʿAṭṭār & 0.52 & 0.84 & 0.64 \\
ʿAmaq Bukhārī & 0.00 & 0.00 & 0.00 \\
ʿOnsorī & 0.50 & 0.02 & 0.04 \\
Ghāleb Dehlavī & 0.55 & 0.27 & 0.36 \\
Farrokhī Sīstānī & 0.49 & 0.50 & 0.49 \\
Forūghī Bastāmī & 0.56 & 0.49 & 0.52 \\
Faṣiḥī Heravī & 0.45 & 0.09 & 0.15 \\
Qāsem Anvār & 0.72 & 0.78 & 0.75 \\
Qodsī Mashhadī & 0.49 & 0.30 & 0.37 \\
Qaṭrān Tabrīzī & 0.62 & 0.60 & 0.61 \\
Qavāmī Rāzī & 0.25 & 0.25 & 0.25 \\
Labībī & 0.00 & 0.00 & 0.00 \\
Majd-e Hamgar & 0.17 & 0.03 & 0.06 \\
Mujīr al-Dīn Baylaqānī & 0.00 & 0.00 & 0.00 \\
Masʿūd Saʿd Salmān & 0.30 & 0.50 & 0.38 \\
Moshtāq Eṣfahānī & 0.33 & 0.07 & 0.11 \\
Malek al-Shoʿarāʾ Bahār & 0.49 & 0.43 & 0.46 \\
Manūchihrī & 1.00 & 0.18 & 0.31 \\
Mahastī Ganjavī & 0.00 & 0.00 & 0.00 \\
Mawlānā & 0.60 & 0.83 & 0.70 \\
Mīrdāmād & 1.00 & 0.03 & 0.05 \\
Mīrzā Ḥabīb Khorāsānī & 0.50 & 0.02 & 0.04 \\
Mīrzādeh ʿEshqī & 0.67 & 0.55 & 0.60 \\
Nāṣer Bukhārī & 0.56 & 0.19 & 0.28 \\
Nāṣir-i Khusraw & 0.69 & 0.96 & 0.81 \\
Najm al-Dīn Rāzī & 0.00 & 0.00 & 0.00 \\
Neshāṭ Eṣfahānī & 0.00 & 0.00 & 0.00 \\
Niẓāmī & 0.79 & 0.87 & 0.82 \\
Halālī Jughṭāʾī & 0.80 & 0.56 & 0.66 \\
Homām Tabrīzī & 0.00 & 0.00 & 0.00 \\
Vahshī Bāfaqī & 0.30 & 0.05 & 0.08 \\
Vaṭvāṭ & 0.61 & 0.61 & 0.61 \\
Kasāʾī & 0.00 & 0.00 & 0.00 \\
Kamāl Khojandī & 0.52 & 0.26 & 0.35 \\
Kamāl al-Dīn Esmāʿīl & 0.27 & 0.41 & 0.32 \\
\bottomrule
\end{longtable}

\end{document}